\documentclass[sigconf, arxiv, authorversion]{acmart}
\setcopyright{none}

\copyrightyear{2025}
\acmYear{2025}
\setcopyright{acmlicensed}
\acmConference[MM '25]{Proceedings of the 33rd
ACM International Conference on Multimedia}{October 27--31, 2025}{Dublin,
Ireland}
\acmBooktitle{Proceedings of the 33rd ACM International Conference on
Multimedia (MM '25), October 27--31, 2025, Dublin, Ireland}
\acmDOI{10.1145/3746027.3754548}
\acmISBN{979-8-4007-2035-2/2025/10}

\usepackage{makecell}
\usepackage{multirow}
\usepackage{colortbl}
\newcommand{\best}{\cellcolor{red!40}}
\newcommand{\sbest}{\cellcolor{orange!50}}
\newcommand{\tbest}{\cellcolor{yellow!50}}
\AtBeginDocument{%
  }

\begin{document}

\title{TSGS: Improving Gaussian Splatting for Transparent Surface Reconstruction via Normal and De-lighting Priors}


\author{Mingwei Li}
\email{mingweili@zju.edu.cn}
\authornote{Also with Zhongguancun Academy.}
\affiliation{%
  \institution{Zhejiang University}
  % \department{College of Computer Science and Technology}
  \city{Hangzhou}
  \state{Zhejiang}
  \country{China}
}
% \additionalaffiliation{%
%   \institution{Zhongguancun Academy}
%   \city{Beijing}
%   \country{China}
% }

\author{Pu Pang}
\email{fankewen@stu.xjtu.edu.cn}
\authornotemark[1]
\affiliation{%
  \institution{Xi'an Jiaotong University}
  \city{Xi'an}
  \state{Shaanxi}
  \country{China}
}
% \additionalaffiliation{%
%   \institution{Zhongguancun Academy}
%   \city{Beijing}
%   \country{China}
% }

\author{Hehe Fan}
\email{hehefan@zju.edu.cn}
\affiliation{%
  \institution{Zhejiang University}
  % \department{College of Computer Science and Technology}
  \city{Hangzhou}
  \state{Zhejiang}
  \country{China}
}

\author{Hua Huang}
\email{huahuang@bnu.edu.cn}
\affiliation{%
  \institution{Beijing Normal University}
  \city{Beijing}
  \country{China}
}

\author{Yi Yang}
\authornote{Corresponding author.}
\email{yangyics@zju.edu.cn}
\affiliation{%
  \institution{Zhejiang University}
  % \department{College of Computer Science and Technology}
  \city{Hangzhou}
  \state{Zhejiang}
  \country{China}
}

\renewcommand{\shortauthors}{Mingwei Li, Pu Pang, Hehe Fan, Hua Huang, and Yi Yang}

\begin{abstract}
Reconstructing transparent surfaces is essential for tasks such as robotic manipulation in labs, yet it poses a significant challenge for 3D reconstruction techniques like 3D Gaussian Splatting (3DGS). These methods often encounter a transparency-depth dilemma, where the pursuit of photorealistic rendering through standard $\alpha$-blending undermines geometric precision, resulting in considerable depth estimation errors for transparent materials. 
To address this issue, we introduce \textbf{Transparent Surface Gaussian Splatting (TSGS)}, a new framework that separates geometry learning from appearance refinement. In the geometry learning stage, TSGS focuses on geometry by using specular-suppressed inputs to accurately represent surfaces. 
In the second stage, TSGS improves visual fidelity through anisotropic specular modeling, crucially maintaining the established opacity to ensure geometric accuracy. To enhance depth inference, TSGS employs a first-surface depth extraction method. This technique uses a sliding window over $\alpha$-blending weights to pinpoint the most likely surface location and calculates a robust weighted average depth. 
To evaluate the transparent surface reconstruction task under realistic conditions, we collect a \textbf{TransLab} dataset that includes complex transparent laboratory glassware.  
Extensive experiments on TransLab show that TSGS achieves accurate geometric reconstruction and realistic rendering of transparent objects simultaneously within the efficient 3DGS framework. 
Specifically, TSGS significantly surpasses current leading methods, achieving a 37.3\% reduction in chamfer distance and an 8.0\% improvement in F1 score compared to the top baseline. 
Additionally, TSGS maintains high-quality novel view synthesis, evidenced by a 0.41dB gain in PSNR, demonstrating that TSGS overcomes the transparency-depth dilemma. 
The code and dataset are available at \url{https://longxiang-ai.github.io/TSGS/}.
\end{abstract}




\begin{teaserfigure}
    \centering
    \includegraphics[width=\textwidth]{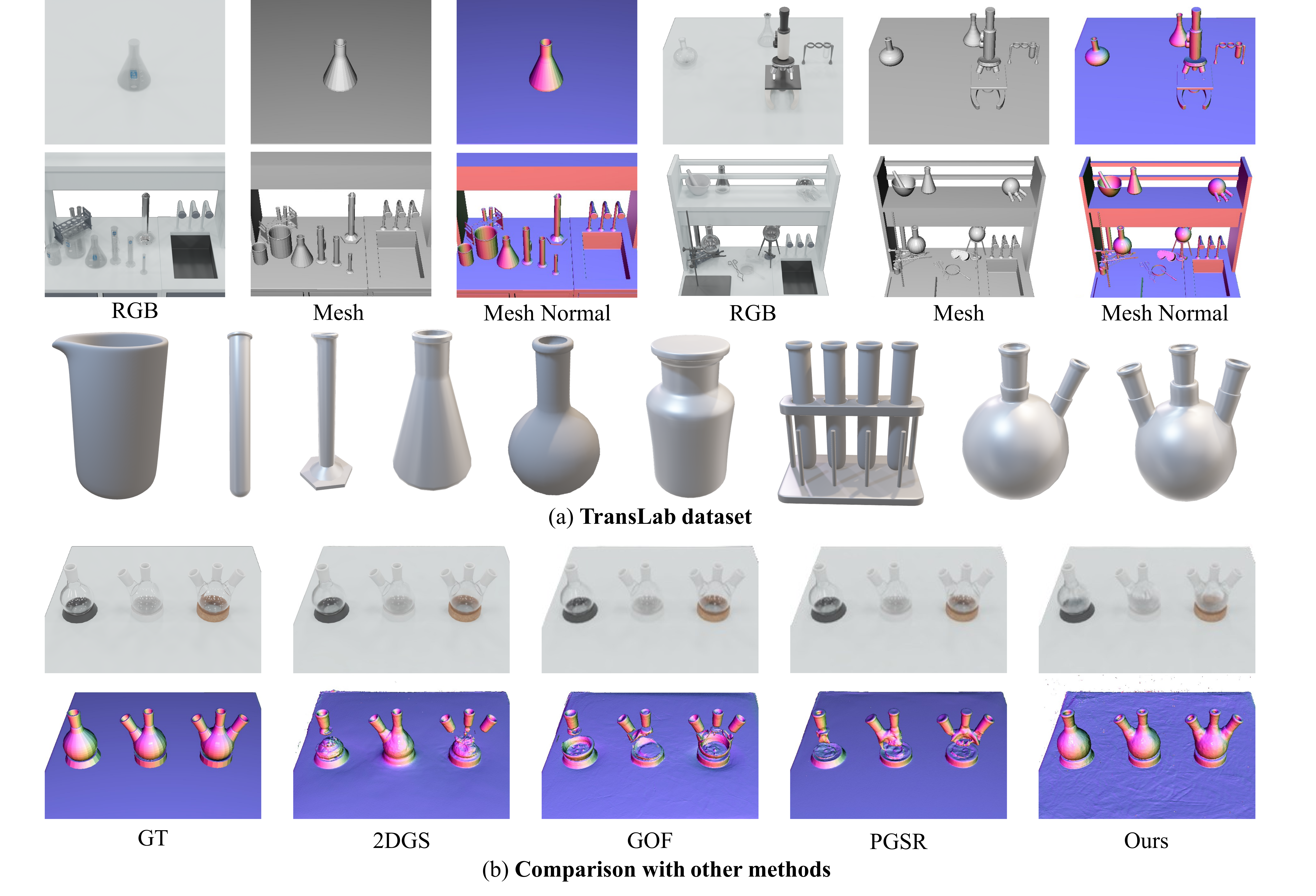}
    \vspace{-1.0cm}
    \caption{ 
    (a) We introduce \textbf{TransLab}, a novel dataset specifically designed for evaluating transparent object reconstruction. TransLab comprises 8 diverse, high-resolution 360° scenes in laboratory settings. It features a wide array of challenging transparent glassware, including test tubes, beakers, volumetric flasks etc.
    (b) Comparative results on TransLab demonstrate the superior capability of our proposed \textbf{TSGS} method in accurately reconstructing geometry and appearance compared to prior state-of-the-art approaches.}
    \Description{Teaser image for the paper.}
    \label{fig:teaser}
    \vspace{-0.2cm}
\end{teaserfigure}
\maketitle

\section{Introduction}
\label{sec:intro}
Novel view synthesis and geometry reconstruction are key tasks in computer vision, essential for AR/VR~\cite{deng2022fov,ye2021superplane}, 3D content creation~\cite{tang2023dreamgaussian,poole2022dreamfusion, gao2024luminat2x, chen2024meshanything}, and autonomous driving. 
These applications require high realism and detailed geometry. Techniques like Neural Radiance Fields (NeRF)~\cite{mildenhall2021nerf, Ye2023IntrinsicNeRF, huang2024nerfdet++, ming2022idf} have advanced high-fidelity results~\cite{muller2022instant,barron2021mip,barron2023zip, wang2021neus,li2023neuralangelo}. Recently, 3D Gaussian Splatting (3DGS)~\cite{kerbl20233d} has emerged as a significant advancement by using explicit 3D Gaussians and an efficient, differentiable rasterizer to enable fast training and real-time rendering. However, 3DGS still struggles with accurately reconstructing challenging materials, especially transparent surfaces.

This limitation in reconstructing transparent surfaces is crucial, especially for applications like robotic laboratory systems that require millimeter-precise manipulation of glassware, such as beakers and test tubes. The main issue with 3DGS is its inability to accurately handle transparent surfaces. This is due to a fundamental conflict: the model is optimized primarily for visual appearance, not for accurately deriving depth using standard $\alpha$-blending techniques. We refer to this problem as the \textbf{transparency-depth dilemma}.

\begin{figure}[t!]
	\centering
	\includegraphics[width=\linewidth]{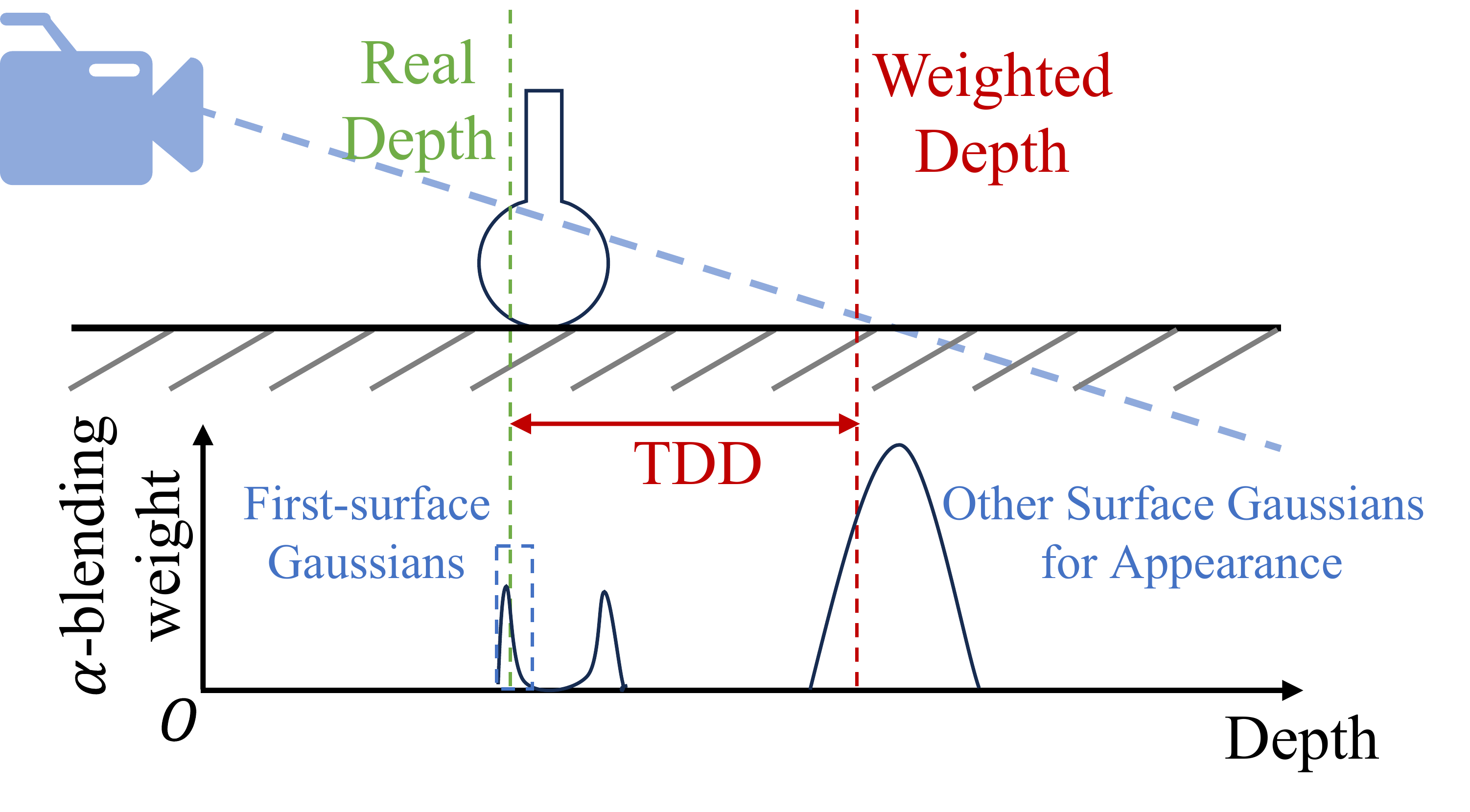}
	% \vspace{-5cm}
	\caption{The transparency-depth dilemma in Gaussian Splatting (\S\ref{sec:intro}): Previous GS-based methods~\cite{gaussianpro,gaussianshader} typically derive depth maps using standard $\alpha$-blending (Eq.~\ref{eq:render_rgb_depth_standard}, \ref{eq:unbiased_depth}), averaging depths of Gaussians along the ray weighted by their appearance-optimized weights. For transparent objects, this blending inherently mixes contributions from the first surface and transmitted background details, failing to accurately localize the true first-surface depth.}
	% \vspace{-7mm}
	\label{fig:td-dilemma}
\end{figure}

Specifically, as shown in Fig.~\ref{fig:td-dilemma}, traditional 3D Gaussian Splatting methods~\cite{Huang2DGS2024, chen2024pgsr} render depth using $\alpha$-blending, where depth values are derived from weighted averages based on the opacity of contributing Gaussians. However, this opacity is mainly optimized for visual appearance rather than accurate depth measurement. While effective for opaque objects—where high-opacity Gaussians cluster near the surface—it fails for transparent materials, giving rise to the transparency-depth dilemma.

The main problem stems from the conflicting goals of modeling appearance and achieving geometric accuracy. In 3DGS, to depict visual transparency and show objects behind a surface, the model reduces the opacity of Gaussians that represent the transparent surface. Consequently, when these appearance-optimized opacities are used to weight depth rendering, the actual transparent surface’s contribution is minimized. This causes deeper background Gaussians to disproportionately affect the weighted average, leading to rendered depths that are systematically greater than the actual physical depths. This results in depth maps that inaccurately depict transparent regions as being farther away than they really are, compromising precise 3D reconstruction.

To address the transparency-depth dilemma, a simple solution might be to abandon $\alpha$-blending and record the depth where a ray first intersects any Gaussian, known as ``nearest depth''. However, this method is flawed because it depends on the properties of individual Gaussian primitives, which do not have stable geometric meaning alone—only the combined effect of Gaussian clusters creates stable surfaces in 3DGS. This leads to two problems. 1) Systematic underestimation bias: In 3DGS, Gaussians are distributed around the true surface. While the surface depth is derived from the collective Gaussian cluster, using the nearest intersection depth typically captures Gaussians closer to the camera than the actual surface. This consistently results in depth values that are smaller (closer) than the true physical depth. 2) Instability and noise artifacts: Since optimization focuses on overall appearance, the precise location and shape of any single Gaussian are not strictly controlled. Therefore, the specific Gaussian that causes the nearest intersection can vary greatly, particularly near object boundaries and thin structures. This leads to noisy and unreliable depth estimations.

To address the outlined challenges, we propose a novel two-stage training framework that separates geometry and appearance optimization, complemented by a first-surface extraction method. 
 In the first stage, we use normal maps and de-lighted images, which render transparent regions as darker and less transparent, to create geometrically accurate Gaussian distributions. 
In the second stage, we fix the opacity parameters of the Gaussians and enhance the model's ability to depict complex transparent appearances using anisotropic spherical Gaussian (ASG) components. 
This approach maintains the accurate geometry established in the first stage while enabling photorealistic rendering of transparency.
At inference time, we introduce a physically principled first-surface extraction technique that accurately pinpoints the initial surface intersection for each ray. This method overcomes the transparency-depth dilemma and avoids the noise artifacts typical of simpler nearest-depth approaches. 

Widely used datasets~\cite{jensen2014large,knapitsch2017tanks,barron2022mip} lack specialized focus on transparent objects with complex geometries like laboratory glassware, yet the growing field of embodied AI for automated labs demands precise reconstruction and manipulation capabilities for these transparent items. To address this gap, we collect the TransLab dataset specifically designed to evaluate transparent surface reconstruction. This dataset includes 8 different 360° scenes of common laboratory environments with transparent equipment like test tubes, beakers, and flasks. 
Our experimental results show significant improvements over existing methods, especially the strong PGSR baseline. With TransLab, our TSGS method reduced the chamber distance by 37.3\% and improved the F1 score for geometric precision by 8.0\%. It also enhanced visual quality, increasing PSNR by +0.41dB. Additionally, TSGS achieved top geometric accuracy on the opaque DTU dataset~\cite{jensen2014large}, demonstrating the robustness and versatility of our approach. 

Overall, our contributions can be summarized as follows:
\vspace{-3mm}
\begin{itemize}
    \item A two-stage training strategy that initially sets up accurate geometry through specular-decoupled learning, followed by appearance refinement while maintaining geometric integrity. This approach ensures robust reconstruction of transparent objects. 
    \item A first-surface depth extraction technique that achieves high accuracy in transparent surface reconstruction through a maximum-weight window aggregation algorithm.
    \item A TransLab dataset that features transparent laboratory equipment in realistic environments. Extensive experiments  show significant improvements of our method over current state-of-the-art methods. 
\end{itemize}
\vspace{-1.5em}
\section{Related Work}
\label{sec:related}
\noindent\textbf{Traditional Surface Reconstruction.}
Early methods operated on unstructured 3D point clouds, utilizing techniques such as Poisson reconstruction~\cite{kazhdan2006poisson}, Delaunay triangulation~\cite{boissonnat1984geometric}, Alpha shapes~\cite{edelsbrunner1994three}, and MLS~\cite{levin2004mesh}. Alternatively, volumetric methods represented geometry via discrete 3D fields, as seen in Marching Cubes~\cite{lorensen1987marching}, level-set methods~\cite{zhao2001fast}, and signed distance fields (SDFs)~\cite{curless1996volumetric}. Multi-view geometry techniques leveraged camera information to recover 3D structure, including SfM~\cite{schonberger2016structure}, MVS~\cite{furukawa2015multi}, and Shape-from-Shading~\cite{zhang1999shape}. Many multi-view pipelines involved intermediate representations like point clouds~\cite{lhuillier2005quasi}, volumes~\cite{kutulakos2000theory}, or depth maps~\cite{schoenberger2016mvs}, establishing dense correspondences via patch-based matching~\cite{barnes2009patchmatch} before generating surfaces through triangulation~\cite{cazals2006delaunay} or implicit fitting~\cite{kazhdan2006poisson}. Despite widespread adoption, these traditional methods are sensitive to noise and struggle with transparent or reflective surfaces that violate photometric assumptions. Recent works improve robustness by integrating deep learning into matching~\cite{wang2021patchmatchnet, sarlin2019coarse}. In this work, we assume known poses (e.g., via SLAM~\cite{campos2021orb, chen2021rnin, chen2022vip} or SfM~\cite{Moulon2012, wu2013towards}) and focus on surface reconstruction under fixed camera parameters.
\begin{figure*}[t!]
    \centering
    \includegraphics[width=\textwidth]{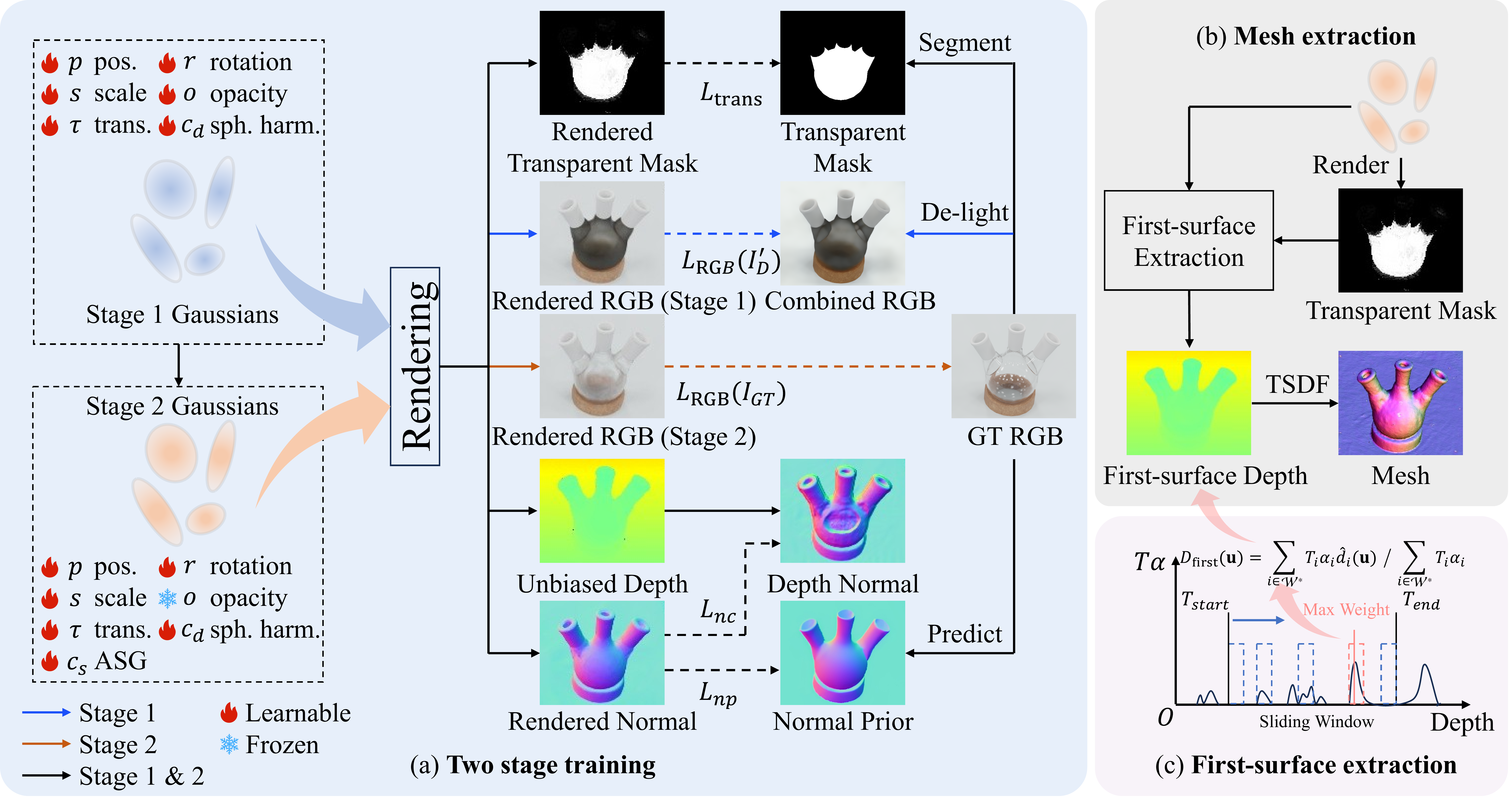}
    \vspace{-0.8cm}
    \caption{Pipeline of TSGS. (a) The two-stage training process. In Stage 1, 3D Gaussians are optimized using geometric priors and de-lighted image inputs (\S\ref{subsec:stage1}). In Stage 2, we fix the opacity parameters and optimize the 3DGS using GT images with specular highlights (\S\ref{subsec:stage2}). (b) The inference process utilizes the trained model to extract the first-surface depth map, which is then used for mesh reconstruction (\S\ref{subsec:first-surface-depth-extraction}). (c) Details of the first-surface depth extraction module, which employs a sliding window to find the maximum-weight window along the ray and computes depth via weighted averaging within this window.}
    %\vspace{-0.5cm}
    \label{fig:pipeline}
\end{figure*}

\noindent\textbf{Neural Surface Reconstruction.}
Neural methods have advanced surface reconstruction by leveraging learned priors and differentiable rendering. Early approaches attempted to directly predict 3D shapes from images, representing geometry as point clouds~\cite{Fan:2017:Point, Lin:2018:Learning}, voxels~\cite{Choy:2016:3d, Xie:2019:Pix2vox}, meshes~\cite{Wang:2018:Pixel2mesh, li2020saliency}, or implicit fields~\cite{Park:2019:Deepsdf, Mescheder:2019:Occupancy}. While enabling end-to-end training, these often require extensive 3D supervision and incur high computational costs. A significant breakthrough came with NeRF~\cite{mildenhall2021nerf}, which represent scenes as continuous 5D functions. Subsequent variants like Mip-NeRF~\cite{barron2021mip} and Zip-NeRF~\cite{barron2023zip} improved rendering quality, while others like NeuS~\cite{wang2021neus} and VolSDF~\cite{Yariv:2021:Volume} adapted NeRF specifically for high-fidelity surface reconstruction using SDF-based representations. To address the computational bottlenecks of purely MLP-based models, accelerated representations have been developed, decomposing scenes into structures like points~\cite{xu2022point}, voxels~\cite{Liu:2020:Neural}, or hybrid combinations~\cite{li2022vox, li2023neuralangelo}, thereby improving efficiency without sacrificing reconstruction quality.
Several methods~\cite{Deng:CVPR2024, lyu2020differentiablerefraction-tracing, Wang_2023_ICCV, gao2023transparent, sun2024nu-nerf, zhang2024from, wu2025alphasurf} have been proposed for transparent object reconstruction. These approaches often leverage NeRF-based techniques that model physical effects like refraction through volumetric ray integration, but this leads to lengthy training times and slow rendering speeds.
% Neural Surface Refinement for Modeling Transparent Objects~\cite{Deng:CVPR2024}
% Differentiable Refraction-Tracing for Mesh Reconstruction of Transparent Objects~\cite{lyu2020differentiablerefraction-tracing}
% NEMTO: Neural Environment Matting for Novel View and Relighting Synthesis of Transparent Objects~\cite{Wang_2023_ICCV}
% Transparent Object Reconstruction via Implicit Differentiable Refraction Rendering~\cite{gao2023transparent}
% NU-NeRF: Neural Reconstruction of Nested Transparent Objects with Uncontrolled Capture Environment~\cite{sun2024nu-nerf}
% From transparent to opaque: rethinking neural implicit surfaces with $\alpha$-NeuS~\cite{zhang2024from}
% $\alpha$surf: Implicit surface reconstruction for semi-transparent and thin objects with decoupled geometry and opacity~\cite{wu2025alphasurf}
% 3D Gaussian Ray Tracing: Fast Tracing of Particle Scenes~\cite{loccoz20243dgrt}
% 3DGUT: Enabling Distorted Cameras and Secondary Rays in Gaussian Splatting~\cite{wu20253dgut}

\noindent\textbf{Gaussian Splatting-based Surface Reconstruction.}
A recent paradigm shift occurred with 3DGS~\cite{kerbl20233d}, enabling real-time rendering and fast training via explicit 3D Gaussians. Extracting consistent surfaces from this representation has been an active area of research. Initial work like SuGaR~\cite{guedon2023sugar} focused on mesh extraction through Poisson reconstruction. Subsequent efforts aimed to enhance geometric quality directly within the Gaussian framework like 2DGS~\cite{Huang2DGS2024}, GOF~\cite{yu2024gaussian}, and PGSR~\cite{chen2024pgsr}. While these methods improve reconstruction for many scenarios, accurately capturing \textbf{transparent surfaces} presents specific nuances. Some approaches~\cite{li2025cargsaddressingreflectivetransparent, kim2025transplatsurfaceembeddingguided3d} employ diffusion priors~\cite{ye2024stablenormal, deweizhou2025dreamrenderer} to aid reconstruction, whereas recent studies~\cite{loccoz20243dgrt, wu20253dgut} incorporate ray tracing at the cost of substantial computational overhead.
A key observation is that the standard $\alpha$-blending mechanism, central to 3DGS rendering, creates an inherent challenge for transparent object geometry: the transparency-depth dilemma. 
While optimizing opacity primarily for appearance successfully enhances visual fidelity, it can lead to inaccuracies when directly using the resulting weighted average for depth calculation. Recognizing this conflict, our work specifically targets the transparency-depth dilemma by proposing a novel approach for accurate first-surface depth extraction from the standard, appearance-optimized Gaussians, aiming to improve geometric reconstruction for transparent objects.
\vspace{-2mm}
\section{Method}
\label{sec:method}
We introduce TSGS to tackle the transparency-depth dilemma in transparent surface reconstruction, where photorealistic rendering conflicts with accurate first-surface geometry. As shown in Fig.~\ref{fig:pipeline}, TSGS employs a three-stage strategy:
1) \textbf{Geometry-Focused Learning} (\S\ref{subsec:stage1}): learns accurate geometry using de-lighted images~\cite{ye2024stablenormal} and normal priors to suppress specular interference.
2) Geometry-Preserving \textbf{Appearance Refinement} (\S\ref{subsec:stage2}): refines appearance with original images and Anisotropic Spherical Gaussians (ASG) while preserving geometry.
3) \textbf{First-Surface Depth Extraction} (\S\ref{subsec:first-surface-depth-extraction}): extracts accurate first-surface depth using an accumulated transmittance sliding window algorithm during inference.
This staged approach yields reconstructions that are both geometrically accurate and visually realistic.

\subsection{Preliminary}
\label{subsec:preliminary}
\noindent\textbf{Gaussian Splatting.}
3DGS~\cite{kerbl20233d} represents scenes using a set of 3D Gaussians \(\{\mathcal{G}_i\}\). Each Gaussian \(\mathcal{G}_i\) is parameterized by its center \(\mathbf{p}_i \), opacity \(o_i\), color \(\mathbf{c}_i\), rotation \(\mathbf{r}_i\) , and scale \(\mathbf{s}_i\). These parameters define the 3D Gaussian distribution in world coordinates as $\mathcal{G}_i(\mathbf{x}) = e^{-\frac{1}{2}(\mathbf{x}-\mathbf{p}_i)^T \mathbf{\Sigma}^{-1}(\mathbf{x}-\mathbf{p}_i)}$, where the covariance \(\mathbf{\Sigma}_i\) is derived from \(\mathbf{r}_i\) and \(\mathbf{s}_i\). These 3D Gaussians are then projected onto the 2D image plane. Given the camera extrinsics \(W\) and intrinsics \(K\), the 3D center \(\mathbf{p}_i\) and covariance \(\mathbf{\Sigma}_i\) are projected to 2D screen coordinates \(\mathbf{p}_i^{\prime}\) and covariance \(\boldsymbol{\Sigma}_i^{\prime}\) via \( \mathbf{p}_i^{\prime} = K W[\mathbf{p}_i, 1]^{\top} \) and \( \boldsymbol{\Sigma}_i^{\prime} = J W \mathbf{\Sigma}_i W^{\top} J^{\top} \), where \(J\) is the Jacobian of the affine approximation for the projective transformation. The depth \(z_i\) of the Gaussian center in camera coordinates is obtained via the camera transformation: \([x_i, y_i, z_i, 1]^{\top} = W[\mathbf{p}_i, 1]^{\top}\).

\noindent\textbf{Unbiased Depth Rendering.}
Standard 3DGS renders color \(\mathbf{C}\) for each pixel \(\mathbf{u}\) via \(\alpha\)-blending. Previous methods~\cite{gaussianpro, gaussianshader} typically render the depth map \(D\) by blending the camera-space depth \(z_i\) of each Gaussian using the same \(\alpha\)-blending weights:
\begin{equation}
    \mathbf{C} = \sum_{i \in N} T_i \alpha_i \mathbf{c}_i, \quad
	D_{\text{standard}} = \sum_{i \in N} T_i \alpha_i z_i
	\label{eq:render_rgb_depth_standard} % New label for standard rendering
\end{equation}
Here, the rendering process involves sorting the Gaussians intersecting the pixel ray by its center depth. For each Gaussian \(\mathcal{G}_i\) along the sorted list, \(T_i = \prod_{j=1}^{i-1}(1 - \alpha_j)\) represents the accumulated transmittance, indicating the probability that the ray has passed through the preceding Gaussians from \(1\) to \(i-1\) without being absorbed. The term \(\alpha_i = o_i \mathcal{G}_i(\mathbf{u}|\mathbf{p}_i^{\prime}, \boldsymbol{\Sigma}_i^{\prime})\) is the opacity contribution of the current Gaussian \(\mathcal{G}_i\) at pixel \(\mathbf{u}\). 
% The standard \(\alpha\)-blending procedure then weights the color \(\mathbf{c}_i\) (or depth \(z_i\)) of each Gaussian by its effective contribution \(w_i\), calculated as:
% \begin{equation}
% 	w_i = T_i \alpha_i 
% 	\label{eq:alpha_blending_weight} 
% \end{equation}
% These weights \(w_i\) essentially quantify how much each Gaussian contributes to the final pixel outcome after accounting for occlusion by preceding Gaussians.

To encourage Gaussians to form planar structures, PGSR~\cite{chen2024pgsr} introduces a flatten regularization \(\mathcal{L}_{\text{flatten}}\) during training:
\begin{equation}
    \mathcal{L}_{\text{flatten}} = \sum_{i} \left | \min(s_{i,1}, s_{i,2}, s_{i,3}) \right |_1.
    \label{eq:scale_loss}
\end{equation}
This allows approximating the normal \(\mathbf{n}_i\) of the Gaussian $\mathcal{G}_i$ from the axis of the smallest scale component. Building on this, PGSR~\cite{chen2024pgsr} introduces an unbiased depth rendering method as an alternative to \(D_{\text{standard}}\). First, it renders both a world-space normal map \(\mathbf{N}_w\) (blending Gaussian normals \(\mathbf{n}_i\)) and a distance map \(\mathcal{D}\) (blending per-Gaussian distances \(d_i\) from the camera center to the Gaussian plane):
\begin{equation}
    \mathbf{N}_w = \sum_{i \in N} T_i \alpha_i \mathbf{n}_i, \quad
    \mathcal{D} = \sum_{i \in N} T_i \alpha_i d_i.
    \label{eq:render_normal_distance_pgsr} % Combined label
\end{equation}
% Second, the unbiased depth map \(D_{\text{unbiased}}(p)\) for a pixel \(p\) is derived by the rendered distance \(\mathcal{D}(p)\), the rendered world normal \(\mathbf{N}_w(p)\), and the ray direction \(\mathbf{v}_p\) passing through the pixel. The ray direction \(\mathbf{v}_p\) is obtained from camera intrinsics \(K\) as \(\mathbf{v}_p = K^{-1}\tilde{p}\), where \(\tilde{p}\) is the homogeneous coordinate:
Second, the unbiased depth map \(D_{\text{unbiased}}(\mathbf{u})\) for a pixel \(\mathbf{u}\) is derived by the ray direction \(\mathbf{v}_\mathbf{u}\), the distance \(\mathcal{D}(\mathbf{u})\) from the plane and the plane normal \(\mathbf{N}_w(\mathbf{u})\) in the world coordinate system:
\begin{equation}
    D_{\text{unbiased}}(\mathbf{u}) = \frac{\mathcal{D}(\mathbf{u})}{\mathbf{N}_w(\mathbf{u}) \cdot \mathbf{v}_\mathbf{u}}.
    \label{eq:unbiased_depth}
\end{equation}
While the unbiased depth \(D_{\text{unbiased}}\) offers improvements over the standard blended depth \(D_{\text{standard}}\), it ultimately derives depth from quantities (\(\mathcal{D}\) and \(\mathbf{N}_w\)) obtained via standard $\alpha$-blending. This process inherently weights contributions from Gaussians in all surfaces intersected along the ray, using weights optimized primarily for appearance. Such averaging across depth makes it challenging to accurately isolate the first surface for transparent objects, reflecting the core transparency-depth dilemma and motivating our alternative first-surface extraction approach detailed in \S\ref{subsec:first-surface-depth-extraction}.

\subsection{Stage 1: Geometry-Focused Learning}
\label{subsec:stage1}

\noindent\textbf{Pre-processing.} 
% 怎么获得normal map, delighted image, transparent mask
Reconstructing the geometry of transparent objects solely from RGB images is inherently \textbf{ill-posed} due to the ambiguity caused by complex light condition. To address this, we leverage strong geometric priors obtained from large-scale diffusion models. Specifically, we utilize normal maps $\{N_{p}\}$ generated by StableNormal~\cite{ye2024stablenormal} and de-lighted image sets $\{I_{D}\}$ pre-processed by StableDelight~\cite{ye2024stablenormal}. To identify the transparent regions, we employ the Grounded-SAM model~\cite{ravi2024sam2segmentimages, liu2023grounding, ren2024grounding, ren2024grounded, kirillov2023segany, jiang2024trex2} to segment the transparent parts across the dataset, obtaining transparency masks $\{M_{T}\}$. Using the transparency mask $M_T$, we blend the de-lighted image $I_D$ with the original ground truth image $I_{GT}$ to create a hybrid image $I'_D$, where only the transparent regions identified by $M_T$ are de-lighted (i.e., specular components are suppressed):
\begin{equation}
    I'_D = M_T \odot I_D + (1 - M_T) \odot I_{GT}
    \label{eq:hybrid_delighted_image}
\end{equation}
% These generated normal maps \(N_p\), hybrid delighted images \(I'_D\), and transparency masks \(M_T\) serve as strong priors. \(N_p\) and \(I'_D\) primarily guide geometry learning by simplifying appearance.

\noindent\textbf{Transparency Attribute Learning.}
To enhance the model's understanding of transparency, we introduce a learnable attribute $\tau_i \in [0, 1]$ for each Gaussian $\mathcal{G}_i$, indicating if it represents a transparent ($\tau_i \ge 0.5$) or opaque ($\tau_i < 0.5$) region. To supervise this attribute, we set a transparency threshold $\theta_T$ and render a predicted transparency mask by selecting the transparency attribute of the Gaussian whose accumulated transmittance $T_i$ is just above this threshold: $\hat{M}_T(\mathbf{u}) = \tau_j$, where $j = \text{argmax}_i \{T_i \ge \theta_T\}$. We then enforce consistency between $\hat{M}_T$ and the ground truth mask $M_T$ using a Binary Cross-Entropy (BCE) loss:
\begin{equation}
    \mathcal{L}_{\text{trans}} = \text{BCE}(\hat{M}_T, M_T).
    \label{eq:transparency_loss} % Added label here
\end{equation}
% This loss guides the model to assign appropriate $\tau_i$ values to Gaussians contributing to transparent regions.

\noindent\textbf{Appearance Modeling.} We primarily focus on learning accurate geometry rather than complex appearance details like specular highlights in Stage 1. Therefore, the appearance \(\mathbf{c}_{i, 1}\) for Stage 1 is represented solely by its diffuse color component \(\mathbf{c}_{i, d}\), modeled using standard Spherical Harmonics (SH)~\cite{kerbl20233d}. This diffuse component is learned by optimizing the appearance loss \(\mathcal{L}_{\text{s1}}\) against the hybrid de-lighted images \(I'_D\):
\begin{equation}
    \mathcal{L}_{\text{s1}} =  \mathcal{L}_{\text{RGB}}(\hat{I}, I'_D) =  (1-\lambda_r) \left| I'_D - \hat{I} \right|_1 + \lambda_r (1 - \text{SSIM} (I'_D, \hat{I}))
    \label{eq:stage1_rgb_loss}
\end{equation}
where $\lambda_r = 0.2$ balances the L1 and SSIM contributions.

\noindent\textbf{Normal Regularization.} As highlighted in the pre-processing (\S\ref{subsec:stage1}), reconstructing transparent object geometry solely from RGB images is an ill-posed problem. To mitigate ambiguity and provide strong geometric guidance, especially in textureless or highly reflective areas, we incorporate normal supervision into training.

\begin{figure}[tp]
    \centering
    \includegraphics[width=\linewidth]{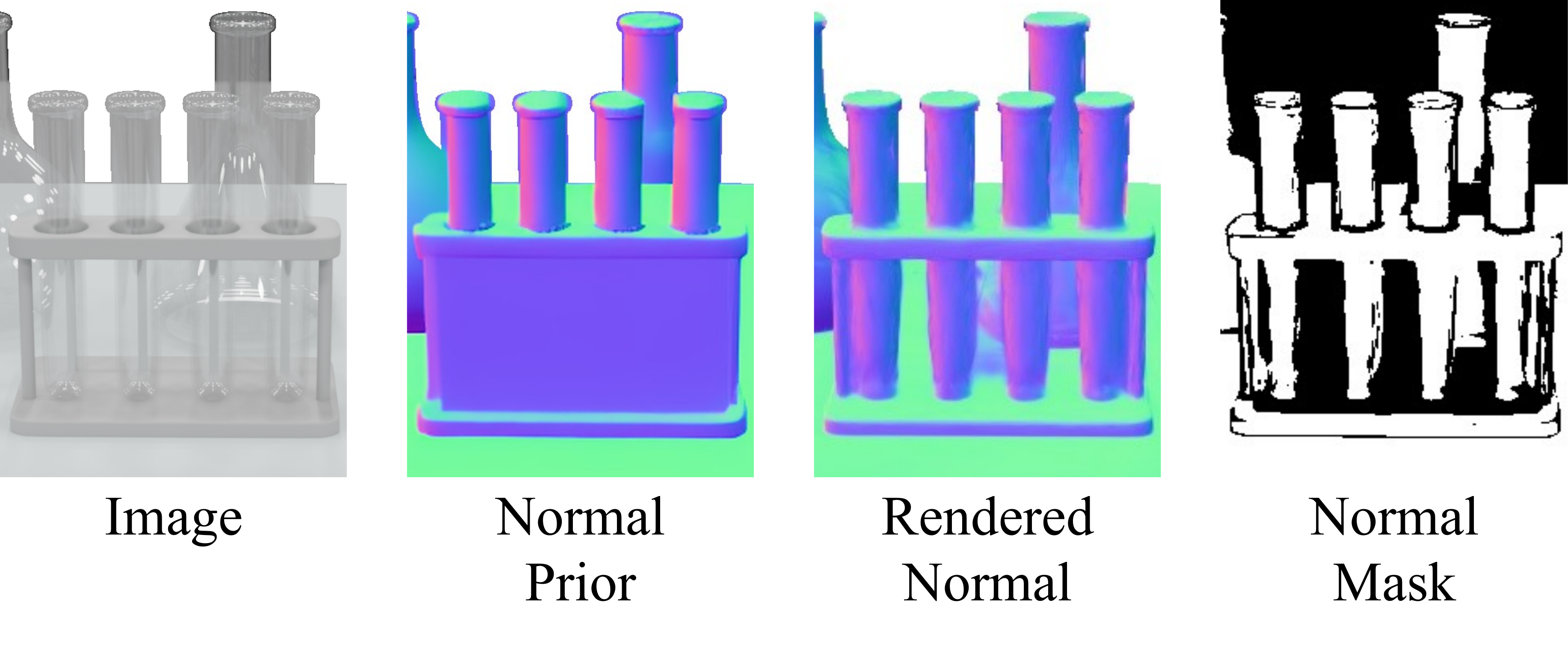}
    \vspace{-1.0cm}
    \caption{Normal prior mask. Normal priors can be inaccurate for challenging views, potentially misleading the rendered normal. Our normal mask filters out these unreliable priors during loss computation, enhancing the robustness of our normal rendering.  (\S\ref{subsec:stage1})}
    \vspace{-0.4cm}
    \label{fig:normal_loss}
\end{figure}
\noindent\textit{Normal Prior Loss ($\mathcal{L}_{\text{np}}$):} To leverage external geometric information, we enforce consistency between the rendered normals \(\mathbf{N}_r\) and the prior normal map \(N_p\) obtained from StableNormal~\cite{ye2024stablenormal}. However, predicted priors \(N_p\) can sometimes be inaccurate due to complex lights. To prevent incorrect priors from degrading the learned geometry, we employ a robust masking strategy, as illustrated in Fig.~\ref{fig:normal_loss}. The loss is calculated as:
\begin{equation}
	\mathcal{L}_{\text{np}} = \sum M_{n} \cdot (1 - \mathbf{N}_{p}^T \mathbf{N}_{r}).
	\label{eq:normal_prior_loss}
\end{equation}
Here, the mask \(M_{n} = (\mathbf{N}_{p}^T \mathbf{N}_{r} \ge \theta_n)\), with a threshold \(\theta_n\), effectively ignores the prior loss contribution for pixels where the rendered normal \(\mathbf{N}_{r}\) significantly deviates from the prior \(\mathbf{N}_{p}\).

\noindent\textit{Depth-Normal Consistency Loss ($\mathcal{L}_{\text{nc}}$):} Following previous practices~\cite{barron2021mipnerf,kerbl20233d}, this loss encourages internal geometric consistency by enforcing agreement between normals computed from the depth map $\mathbf{N}_d$ and the directly rendered normals $\mathbf{N}_r$:
\begin{equation}
	\mathcal{L}_{\text{nc}} = \sum (1 - \mathbf{N}_{d}^T \mathbf{N}_{r}).
	\label{eq:normal_consistency_loss}
\end{equation}
Thus, the normal regularization term is defined as:
\begin{equation}
    \mathcal{L}_{n} = \mathcal{L}_{\text{np}} + \mathcal{L}_{\text{nc}}.
    \label{eq:combined_normal_loss}
\end{equation}

\noindent\textbf{Total Loss for Stage 1.} The final training objective for Stage 1, \(\mathcal{L}_{\text{total, s1}}\), combines the stage-specific appearance loss (\(\mathcal{L}_{\text{s1}}\)) with the transparency loss (\(\mathcal{L}_{\text{trans}}\)), the combined normal loss (\(\mathcal{L}_{n}\)), and the flatten regularization loss \(\mathcal{L}_{\text{flatten}}\) (defined in Eq.~\ref{eq:scale_loss}):
\begin{equation}
	\mathcal{L}_{\text{total, s1}} = \mathcal{L}_{\text{s1}} + \lambda_{t}\mathcal{L}_{\text{trans}} + \lambda_{n}\mathcal{L}_{n} + \lambda_{f}\mathcal{L}_{\text{flatten}}
	\label{eq:total_loss_stage1} % New label for Stage 1 total loss
\end{equation}
where $\lambda_t=0.1$, $\lambda_n=0.1$, and $\lambda_f=100$ are hyperparameters balancing the contributions of the respective regularization terms during this geometry-focused stage.

\subsection{Stage 2: Appearance Refinement}
\label{subsec:stage2}
Having established a reliable geometry foundation in Stage 1, our focus in Stage 2 shifts to enhancing the appearance quality, particularly capturing the complex specular and transparent effects present in the original images \(I_{GT}\). Instead of using the hybrid de-lighted images \(I'_D\) from Stage 1, we supervise with the original ground truth images \(I_{GT}\). Moreover, to preserve the learned geometric structure represented in the opacity distribution while optimizing for more complex appearance, we \textbf{fix} the opacity parameter $o_i$, preventing incorrect updates driven by the appearance loss against $I_{GT}$.

\noindent\textbf{Appearance Modeling.} To effectively model the strong view-dependency of specular highlights and transmission effects common in transparent objects, we utilize the appearance modeling strategy from Spec-Gaussian~\cite{yang2024spec} to model the specular component. Specifically, we employ anisotropic spherical Gaussians (ASGs) to model the latent feature of specular component $\mathbf{c}_{i,s}$:
\begin{equation}
    \mathbf{c}_{i,s} = \Psi(\kappa_i, \gamma(\mathbf{d}_i), \mathbf{n}_i, -\mathbf{d}_i) 
\end{equation}
where $\Psi$ is a tiny MLP decoder, $\kappa_i$ represents the latent feature derived from $\mathbf{K}$ ASG primitives, $\mathbf{d}_i$ denotes the  unit view direction pointing from the camera to each 3D Gaussian, $\gamma$ is the positional encoding. This strategy enhances the ability to model complex optical phenomena compared to using SH directly for color representation. The full appearance component in Stage 2 is:
\begin{equation}
\mathbf{c}_{i, 2} = \mathbf{c}_{i,d} + \mathbf{c}_{i,s}
\label{eq:appearance_stage2}
\end{equation}
The appearance loss for this stage uses the same L1 and SSIM combination as $\mathcal{L}_{\text{s1}}$ (Eq.~\ref{eq:stage1_rgb_loss}), but focuses on reconstructing the appearance from the original ground truth images $I_{GT}$:
\begin{equation}
    \mathcal{L}_{\text{s2}} = \mathcal{L}_{\text{RGB}}(\hat{I}, I_{GT})
    \label{eq:stage2_rgb_loss}
\end{equation}
where $\hat{I}$ is the rendered image using \(\mathbf{c}_{i, 2}\).

\noindent\textbf{Total Loss for Stage 2.} The training objective for Stage 2, \(\mathcal{L}_{\text{total, s2}}\), combines the appearance loss (\(\mathcal{L}_{\text{s2}}\)) with the same set of regularization terms used in Stage 1:
\begin{equation}
	\mathcal{L}_{\text{total, s2}} = \mathcal{L}_{\text{s2}} + \lambda_{t}\mathcal{L}_{\text{trans}} + \lambda_{n}\mathcal{L}_{n} + \lambda_{f}\mathcal{L}_{\text{flatten}}
	\label{eq:total_loss_stage2}
\end{equation}

\subsection{First-surface Depth Extraction}
\label{subsec:first-surface-depth-extraction}
\begin{figure}[t]
    \centering
    \includegraphics[width=\linewidth]{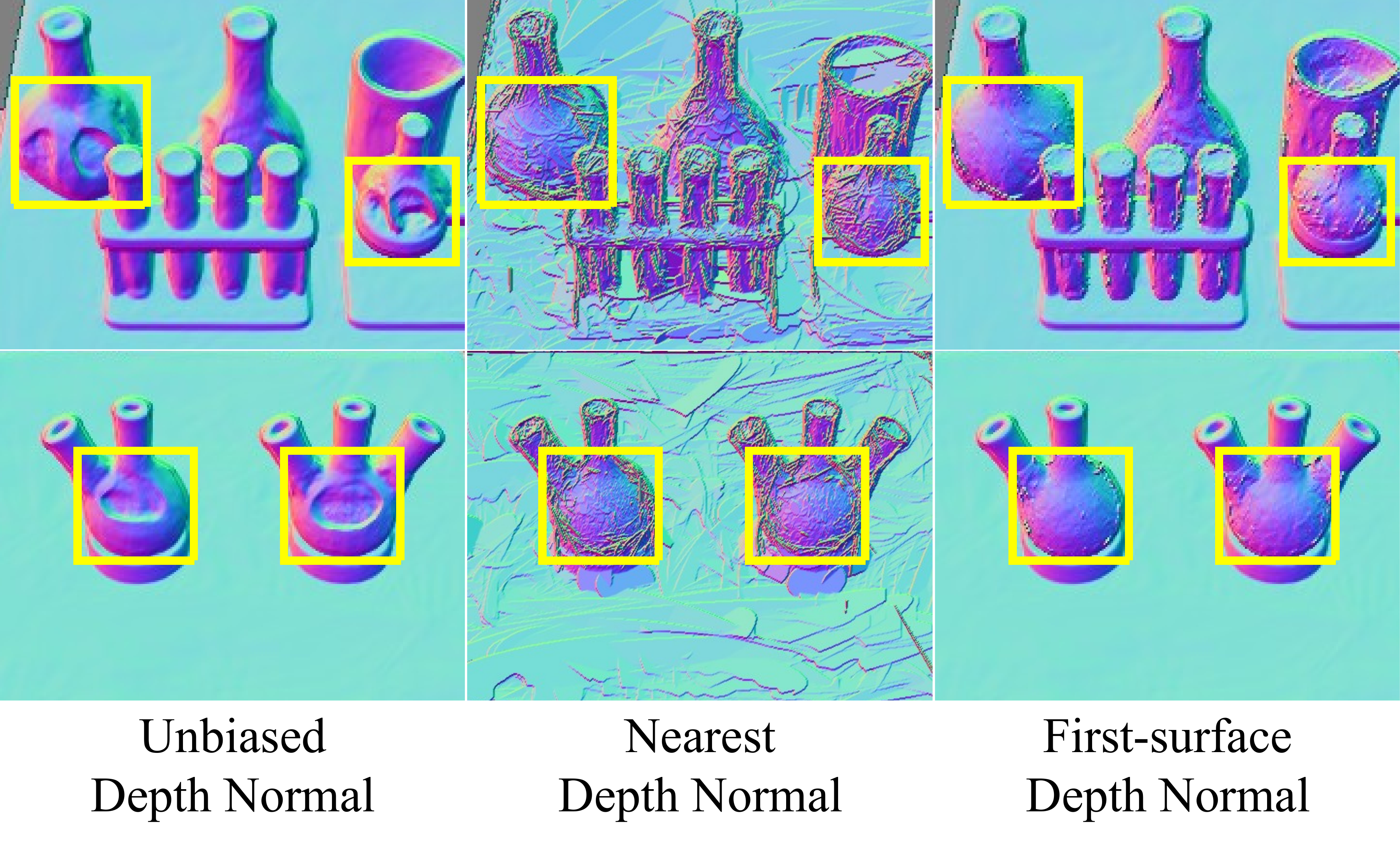}
    \vspace{-0.9cm}
    \caption{Comparison of normal maps derived from different depth extraction methods: unbiased depth~\cite{chen2024pgsr}, nearest depth (\(D_{\text{nearest}}\), \S\ref{subsec:first-surface-depth-extraction}), our first-surface depth (\(D_{\text{first}}\), \S\ref{subsec:first-surface-depth-extraction}). Our method produces significantly smoother and more accurate surface normals. In contrast, nearest depth yields noisy results due to ``floaters'', while normals derived from unbiased depth can exhibit incorrect dips on transparent surfaces due to the underlying $\alpha$-blending across depth.}
    \vspace{-0.7cm}
    \label{fig:different-depth}
\end{figure}

% 1. 先介绍 Gaussian Plane Depth
\noindent\textbf{Gaussian Plane Depth.} 
Inspired by PGSR~\cite{chen2024pgsr}, we utilize the concept of Gaussian plane depth which indicates the intersection depth instead of the Gaussian center depth. Following the principle of unbiased depth rendering~\cite{chen2024pgsr} (Eq.~\ref{eq:unbiased_depth}), we calculate the plane depth \(\hat{d_i}(\mathbf{u})\) for Gaussian \(\mathcal{G}_i\) by the distance \(d_i\), Gaussian normal \(\mathbf{n}_i\) and the ray direction \(\mathbf{v}_\mathbf{u}\):
\begin{equation}
    \hat{d_i}(\mathbf{u}) = \frac{d_i}{\mathbf{n}_i \cdot \mathbf{v}_\mathbf{u}}.
    \label{eq:Gaussian_unbiased_depth} 
\end{equation}
We compute this plane depth \(\hat{d_i}\) for \textit{each individual} Gaussian. This per-Gaussian depth value then serves as the basis for our subsequent depth extraction process, rather than using a weighted average across \textit{all} intersected Gaussians.

% 2. 讨论 Nearest Depth 的局限性
\noindent\textbf{Limitations of Nearest Depth Approach.}
The naive nearest depth discussed in \S\ref{sec:intro} simply takes the minimum of all the plane depths of Gaussian along the ray:
\begin{equation}
	D_{\text{nearest}}(\mathbf{u}) = \min_i\{\hat{d_i}(\mathbf{u}) \}
\end{equation}
% As illustrated in Fig.~\ref{fig:different-depth}, this suffers from systematic underestimation bias and "floaters".
% A seemingly better baseline, which we term "Nearest Plane Depth" (\(D_{\text{nearest}}\)), uses the plane depth \(\hat{d}_k(\mathbf{u})\) of the first intersected Gaussian along the ray (i.e., \(k = \arg\min_i z_i\)). While using \(\hat{d_i}\) is an improvement over \(z_i\), selecting the Gaussian based solely on the minimum \(z_i\) makes this approach highly vulnerable to "floaters" – erroneous Gaussians near the camera – leading to inaccurate depth estimates and noisy derived normals (Fig.~\ref{fig:different-depth}). Both nearest approaches struggle to robustly identify the actual first surface within the appearance-optimized opacity field.
As illustrated in Fig.~\ref{fig:different-depth}, the nearest depth \(D_{\text{nearest}}\) exhibits a systematic underestimation bias and is heavily corrupted by ``floaters'' (erroneous Gaussians). This aligns with our analysis in \S\ref{sec:intro} regarding the limitations of relying on individual Gaussian properties for depth extraction. 
To better address these issues, we need to rethink why previous works struggles to reconstruct transparent surfaces.
% \begin{figure}[tp]
%     \centering
%     \includegraphics[width=0.8\linewidth]{sources/first_plane.png}
%     \caption{First-Surface Depth Extraction. A sliding window identifies the region with maximum summed \(\alpha\)-blending weights (\(T_t \alpha_t\)), and the weighted average depth within this window yields the final \(d_{\text{first}}\). (\S\ref{subsec:first-surface-depth-extraction})}
%     \label{fig:first-surface}
% \end{figure}

\noindent\textbf{Challenge of Transparent Surface Reconstruction.}
The core challenge in transparent surface reconstruction stems from the fact that the opacity distribution required for photorealistic RGB rendering is inherently \textbf{distinct} from the one required for precise first-surface depth extraction. The former often requires lower opacities at the surface to correctly blend foreground and background light transport, while the latter ideally requires opacity concentrated sharply at the true first surface. Our approach, therefore, focuses on extracting a reliable depth-indicative distribution from the appearance-optimized opacity field learned by 3DGS, as illustrated in Fig.~\ref{fig:pipeline} (c).

\noindent\textbf{Accumulated Transmittance Drop.}
For pixels predicted as transparent (\(\hat{M}_T(\mathbf{u}) \ge 0.5\)), our approach leverages the behavior of accumulated transmittance (AT) along the ray.
% Recall that the accumulated transmittance \(T_i = \prod_{j=1}^{i-1}(1 - \alpha_j)\) (used in Eq.~\ref{eq:alpha_blending_weight}) represents the probability that the ray has traveled unimpeded up to Gaussian \(i\). As the rendering process steps through Gaussians sorted by depth, each contributing opacity \(\alpha_i\), this accumulated transmittance \(T_i\) progressively decreases (\(T_{i+1} = T_i(1-\alpha_i)\)). We analyze the pattern of this decrease to identify the first surface.
As formulated in Eq.~\ref{eq:render_rgb_depth_standard}, $T_i$ guadually decreases by the contribution of each Gaussian along the ray. For opaque surfaces, $T_i$ drops sharply from nearly 1 to 0 upon hitting the surface. For transparent surfaces,
% while the first interaction also causes a noticeable drop in AT marking the surface's position, 
this drop is typically less drastic, and $T_i$ does not reach zero due to light penetration. Our objective is thus to robustly locate the depth corresponding to this \textit{first significant drop} in $T_i$.

% To identify the region of primary interaction along the ray, we consider the standard \(\alpha\)-blending weights, \(w_i = T_i \alpha_i\), for each Gaussian \(i\) sorted by depth when rendering pixel \(u\). These weights quantify the contribution of each Gaussian to the final rendered color(as in Eq.~\ref{eq:render_rgb_depth_standard}) and inherently reflect the magnitude of the transmittance drop at each step $i$. They are visualized in Fig.~\ref{fig:pipeline} (c).

\noindent\textbf{Maximum-Weight Window Search.}
We operate under the hypothesis that the region containing the true first surface will exhibit the concentration of Gaussians. To efficiently identify this region, we restrict our search along the ray using thresholds \(T_{\text{start}}\) and \(T_{\text{end}}\). \(T_{\text{start}}\) helps to ignore potential noise or minor interactions near the camera, while \(T_{\text{end}}\) defines a point beyond which the first significant interaction is assumed to have already occurred. We employ a sliding window approach, as illustrated in Fig.~\ref{fig:pipeline} (c), considering only the segment of the ray where the \(T_i \in [T_{\text{end}}, T_{\text{start}}]\). This effectively targets the primary interaction zone where the first surface is most likely located.

Within this relevant ray segment \(\{ \mathcal{G}_i \mid T_{\text{end}} \le T_i \le T_{\text{start}} \}\), we define a window of a fixed size \(\delta t\) and slide it along the sorted candidate Gaussians. For each window position $\mathcal{W}_j$ containing a set of these candidate Gaussians, we calculate the sum of weights within it:
% \begin{equation}
     \( W_{\text{sum}}(j) = \sum_{i \in \mathcal{W}_j} T_i \alpha_i \).
% \end{equation}
We then select the window $\mathcal{W}^*$ that maximizes this sum: $\mathcal{W}^* = \arg\max_j W_{\text{sum}}(j)$.

% 4. 重新组织 First-Surface Depth Calculation
\noindent\textbf{First-surface Depth Calculation.} Unlike the unbiased depth approach (Eq.~\ref{eq:unbiased_depth}) which relies on $\alpha$-blending aggregated quantities (\(\mathcal{D}\) and \(\mathbf{N}_w\)) of all Gaussians along the ray, we compute the first-surface depth \(D_{\text{first}}(\mathbf{u})\) in the identified the maximum-weight window \(\mathcal{W}^*\), which robustly indicates the region of the first surface interaction. We perform a weighted average of the Gaussian plane depths \(\hat{d_i}(\mathbf{u})\) for Gaussians \(\mathcal{G}_i\) falling within this window \(\mathcal{W}^*\):
\begin{equation}
	D_{\text{first}}(\mathbf{u}) = \frac{\sum_{i \in \mathcal{W}^*} T_i \alpha_i \hat{d_i}(\mathbf{u})}{\sum_{i \in \mathcal{W}^*} T_i \alpha_i}
	\label{eq:first_surface_depth_window_avg_unbiased} 
\end{equation}
Crucially, by detecting the transmittance drop to define the window \(\mathcal{W}^*\) and then averaging the plane depths within it, we achieve two advantages. First, restricting the average to \(\mathcal{W}^*\) effectively isolates the first surface's contribution, preventing background details seen through transparency from corrupting the depth estimate and thus addressing the transparency-depth dilemma issue inherent in full $\alpha$-blending approaches (\(D_{\text{standard}}\), \(D_{\text{unbiased}}\)). Second, the weighted averaging within the window provides robustness against the ``floaters'' that plague nearest depth methods (\(D_{\text{nearest}}\)), as these noisy Gaussians typically have low $T_i\alpha_i$ or fall outside \(\mathcal{W}^*\). The entire process is implemented efficiently in parallel using CUDA.
\section{Experiments}
\label{sec:experiments}
\subsection{Benchmark Settings}
\label{subsec:benchmark-settings}
\noindent\textbf{Implementation Details.}
Our training strategy and hyperparameters are generally consistent with 3DGS~\cite{kerbl20233d}. The training iterations for all scenes are set to 30,000, and 15,000 for the two stages, respectively. We adopt the densification strategy of AbsGS~\cite{ye2024absgs}. For mesh extraction, we begin by rendering the first-surface depth for each training view, followed by utilizing the TSDF Fusion algorithm~\cite{newcombe2011kinectfusion} to extract the mesh. For data that does not have transparent objects, we choose not to generate a transparent mask.
All our experiments are conducted on a single NVIDIA A800 GPU.
\begin{table}[t!]
    \centering
    \caption{Quantitative comparison of geometric reconstruction quality across different methods in TransLab dataset. ``Red'', ``Orange'', and ``Yellow'' denote the best, second-best, and third-best results, respectively. Our proposed method outperforms existing mesh reconstruction techniques in terms of reconstruction quality. (\S\ref{subsec:evaluation})}
    \label{tab:geometric_quality}
    \vspace{-0.3cm}
    \resizebox{\columnwidth}{!}{% Add \resizebox here
    \begin{tabular}{l|cccccccc|c}
        \toprule
        \multirow{2}{*}{Method} & \multicolumn{8}{c|}{Scene} & \multirow{2}{*}{Average} \\
        \cmidrule(lr){2-9}
        & Scene 1 & Scene 2 & Scene 3 & Scene 4 & Scene 5 & Scene 6 & Scene 7 & Scene 8 & \\
        \midrule
        \multicolumn{10}{c}{Chamfer Distance $\downarrow$} \\
        \midrule
        SuGaR~\cite{guedon2023sugar}& 37.17 & 10.34 & 8.01 & 6.48 & 8.55 & 46.15 & 14.01 & 7.44 & 17.27 \\
        2DGS~\cite{Huang2DGS2024} & \tbest{2.50} & \tbest{8.63} & \sbest{2.24} & \sbest{2.14} & \tbest{4.80} & \sbest{2.12} & \sbest{2.12} & \tbest{3.86} & \sbest{3.55} \\
        GOF~\cite{yu2024gaussian}  & 4.79 & 10.38 & 2.72 & 3.14 & 10.03 & 18.39 & 9.97 & 4.88 & 8.04 \\
        PGSR~\cite{chen2024pgsr} & \sbest{1.88} & \sbest{5.69} & \tbest{2.51} & \tbest{2.91} & \sbest{2.60} & \tbest{2.27} & \tbest{2.32} & \sbest{3.44} & \tbest{2.95} \\
        Ours & \best{1.68} & \best{2.42} & \best{1.57} & \best{1.60} & \best{1.75} & \best{1.54} & \best{2.05} & \best{2.23} & \best{1.85} \\
        \midrule
        \multicolumn{10}{c}{F1 Score $\uparrow$} \\
        \midrule
        SuGaR~\cite{guedon2023sugar} & 0.25 & 0.49 & 0.59 & 0.56 & 0.57 & 0.26 & 0.34 & 0.46 & 0.44 \\
        2DGS~\cite{Huang2DGS2024} & \tbest{0.91} & 0.55 & \sbest{0.90} & \sbest{0.89} & \tbest{0.69} & \best{0.98} & \best{0.97} & \tbest{0.84} & \tbest{0.84} \\
        GOF~\cite{yu2024gaussian}  & 0.83 & \tbest{0.65} & 0.86 & 0.84 & 0.57 & 0.58 & 0.36 & 0.60 & 0.66 \\
        PGSR~\cite{chen2024pgsr} & \sbest{0.94} & \sbest{0.78} & \tbest{0.86} & \tbest{0.85} & \sbest{0.90} & \tbest{0.93} & \tbest{0.94} & \sbest{0.85} & \sbest{0.88} \\
        Ours & \best{0.97} & \best{0.91} & \best{0.96} & \best{0.95} & \best{0.97} & \best{0.98} & \sbest{0.96} & \best{0.94} & \best{0.95} \\
        \bottomrule
    \end{tabular}% End tabular
    \vspace{-1.0cm}
    } % End \resizebox
\end{table}

\begin{table}[ht]
    \centering
    \vspace{-0.3cm}
    \caption{Quantitative comparison of novel view synthesis across different methods in TransLab dataset. Our method achieves state-of-the-art results in PSNR and SSIM, and shows competitive performance in LPIPS. (\S\ref{subsec:evaluation})}
    \label{tab:image_quality}
    \vspace{-0.4cm}
    \resizebox{\columnwidth}{!}{% Add \resizebox here
    \begin{tabular}{l|cccccccc|c}
        \toprule
        \multirow{2}{*}{Method} & \multicolumn{8}{c|}{Scene} & \multirow{2}{*}{Average} \\
        \cmidrule(lr){2-9}
        & Scene 1 & Scene 2 & Scene 3 & Scene 4 & Scene 5 & Scene 6 & Scene 7 & Scene 8 & \\
        \midrule
        \multicolumn{10}{c}{PSNR $\uparrow$} \\
        \midrule
        SuGaR~\cite{guedon2023sugar} & 23.27 & 23.21 & 22.20 & 21.81 & 22.83 & 23.44 & 24.34 & 23.14 & 23.03 \\
        2DGS~\cite{Huang2DGS2024} & \tbest{38.94} & \tbest{30.12} & \tbest{38.74} & \tbest{37.68} & \tbest{34.60} & \tbest{36.07} & \tbest{41.45} & \tbest{37.68} & \tbest{36.91} \\  
        GOF~\cite{yu2024gaussian} & 24.13 & 24.73 & 23.78 & 23.72 & 23.99 & 24.57 & 23.57 & 24.01 & 24.06 \\
        PGSR~\cite{chen2024pgsr} & \sbest{39.89} & \sbest{36.09} & \sbest{39.76} & \sbest{38.36} & \sbest{34.93} & \sbest{36.80} & \sbest{45.14} & \best{38.37} & \sbest{38.67} \\
        Ours & \best{41.02} & \best{36.58} & \best{40.34} & \best{38.37} & \best{35.41} & \best{37.37} & \best{45.72} & \sbest{37.83} & \best{39.08} \\
        \midrule
        \multicolumn{10}{c}{SSIM $\uparrow$} \\
        \midrule
        SuGaR~\cite{guedon2023sugar} & 0.946 & \tbest{0.941} & 0.933 & 0.921 & 0.935 & 0.944 & \tbest{0.961} & 0.945 & \tbest{0.941} \\
        2DGS~\cite{Huang2DGS2024} & \tbest{0.962} & 0.702 & \tbest{0.953} & \tbest{0.952} & \tbest{0.944} & \tbest{0.960} & \tbest{0.961} & \tbest{0.955} & 0.924 \\
        GOF~\cite{yu2024gaussian} & 0.937 & 0.930 & 0.930 & 0.928 & 0.920 & 0.935 & 0.934 & 0.930 & 0.931 \\
        PGSR~\cite{chen2024pgsr} & \sbest{0.995} & \best{0.981} & \sbest{0.990} & \sbest{0.989} & \sbest{0.976} & \best{0.989} & \best{0.997} & \best{0.988} & \sbest{0.988} \\
        Ours & \best{0.996} & \sbest{0.980} & \best{0.995} & \best{0.992} & \best{0.978} & \sbest{0.985} & \best{0.997} & \sbest{0.987} & \best{0.989} \\
        \midrule
        \multicolumn{10}{c}{LPIPS $\downarrow$} \\
        \midrule
        SuGaR~\cite{guedon2023sugar} & \tbest{0.114} & \tbest{0.133} & 0.144 & 0.157 & \tbest{0.130} & \tbest{0.110} & \tbest{0.081} & \tbest{0.111} & \tbest{0.123} \\
        2DGS~\cite{Huang2DGS2024} & 0.130 & 0.338 & 0.165 & 0.160 & 0.156 & 0.113 & 0.139 & 0.137 & 0.167 \\
        GOF~\cite{yu2024gaussian} & 0.123 & 0.137 & \tbest{0.132} & \tbest{0.133} & 0.144 & 0.118 & 0.130 & 0.131 & 0.131 \\
        PGSR~\cite{chen2024pgsr} & \best{0.010} & \best{0.030} & \sbest{0.020} & \sbest{0.019} & \best{0.027} & \best{0.018} & \sbest{0.007} & \best{0.015} & \best{0.018} \\
        Ours & \best{0.010} & \sbest{0.038} & \best{0.013} & \best{0.015} & \sbest{0.032} & \sbest{0.025} & \best{0.006} & \sbest{0.019} & \sbest{0.020} \\
        \bottomrule
    \end{tabular} % End tabular
    } % End \resizebox
    \vspace{-0.4cm}
\end{table}

\noindent\textbf{Datasets.}
To specifically assess performance on transparent surfaces, we introduce a new specialized dataset called \textbf{TransLab} (Transparent Laboratory). TransLab contains 8 diverse 360° scenes featuring common laboratory environments with various transparent equipment like test tubes, beakers, separatory funnels, safety goggles, flasks (including two-neck and three-neck variants), and petri dishes. This collection comprehensively represents the challenges of reconstructing transparent objects typically found in chemistry laboratories. Additionally, to evaluate general reconstruction quality, we also test our method on 15 object-centric scenes from the standard DTU dataset~\cite{jensen2014large}.

\noindent\textbf{Evaluation Metrics.} 
To evaluate the geometric accuracy of the reconstructed surfaces, our primary focus, we utilize the chamfer distance (CD) and F1 score. We also assess the photorealistic quality of the rendered novel views using standard image synthesis metrics: PSNR, SSIM, and LPIPS~\cite{zhang2018unreasonable}.
\setlength\tabcolsep{0.5em}
\begin{table*}[!ht]
% \captionsetup{font={footnotesize}}
\caption{Quantitative results of chamfer distance (lower is better) on DTU dataset~\cite{jensen2014large}. (\S\ref{subsec:evaluation})} 
\vspace{-0.3cm}
\centering
\resizebox{0.8\textwidth}{!}{%
\begin{tabular}{@{}lccccccccccccccccclc}
\toprule
 \multicolumn{3}{c}{} & 24 & 37 & 40 & 55 & 63 & 65 & 69 & 83 & 97 & 105 & 106 & 110 & 114 & 118 & 122 & & Mean \\ \midrule
 & VolSDF~\cite{Yariv:2021:Volume} & & 1.14 & 1.26 & 0.81 & 0.49 & 1.25 & 0.70 & 0.72 & 1.29 & 1.18 & 0.70 & 0.66 & 1.08 & 0.42 & 0.61 & 0.55 & & 0.86\\
 & NeuS~\cite{wang2021neus} & & 1.00 & 1.37 & 0.93 & 0.43 & 1.10 & 0.65 & 0.57 & 1.48 & 1.09 & 0.83 & \sbest 0.52 & 1.20 & 0.35 & 0.49 & 0.54 & & 0.84\\
 & Neuralangelo~\cite{li2023neuralangelo} & & \tbest 0.37 & \sbest 0.72 & \best 0.35 & \sbest 0.35 & \sbest 0.82 & \best 0.54 & \sbest 0.53 & 1.29 & \tbest 0.97 & 0.73 & 0.47 & \tbest 0.74 & \tbest 0.32 & \sbest 0.41 & \tbest 0.43 & & \tbest 0.61\\ 
 \toprule
 & SuGaR~\cite{guedon2023sugar} && 1.47 & 1.33 & 1.13 & 0.61 & 2.25 & 1.71 & 1.15 & 1.63 & 1.62 & 1.07 & 0.79 & 2.45 & 0.98 & 0.88 & 0.79 & & 1.33\\
 & 2DGS~\cite{Huang2DGS2024} && 0.48 & 0.91 & 0.39 & 0.39 & 1.01 & 0.83 & 0.81 & 1.36 & 1.27 & 0.76 & 0.70 & 1.40 & 0.40 & 0.76 & 0.52 & & 0.80\\
  & GOF~\cite{yu2024gaussian} & & 0.50 & 0.82 & \sbest 0.37 & 0.37 & 1.12 & 0.74 & 0.73 & \tbest 1.18 & 1.29 & \tbest 0.68 & 0.77 & 0.90 & 0.42 & 0.66 & 0.49 & & 0.74\\
 & PGSR~\cite{chen2024pgsr} && \sbest 0.35 & \sbest 0.55 & 0.40 & \best 0.33 & \sbest 0.82 & \sbest 0.55 & \best 0.48 & \sbest 1.10 & \sbest 0.70 & \sbest 0.60 & \sbest 0.52 & \sbest 0.68 & \best 0.30 & \tbest 0.45 & \sbest 0.37 & & \sbest 0.55
 \\
 & Ours && \best 0.34 & \best 0.53 & \sbest 0.37 & \tbest 0.41 & \best 0.79 & \sbest 0.55 & \best 0.48 & \best 1.05 & \best 0.62 & \best 0.59 & \best 0.42 & \best 0.49 & \sbest 0.31 & \best 0.37 & \best 0.35 & & \best 0.51
 \\
 \toprule
\end{tabular}
}
\label{tab:dtu_result}
\vspace{-0.4cm}
\end{table*}
\begin{table}[t]
    \centering
    \caption{\small \textbf{Ablation study.} We evaluate the effectiveness of components in our method on both geometry and appearance metrics. (\S\ref{subsec:ablation-study}) 
% \textit{w/o Normal Prior} and \textit{ w/o Delight Prior} are introduced in Stage 1 to guide geometry reconstruction under transparent and specular conditions.
% \textit{w/o First-Surface Depth} provides physics-based depth extraction to handle transparency.
% \textit{w/o Anisotropic Spherical Gaussians (ASG)} are used in Stage 2 to model view-dependent appearance.
% \textit{w/o Fixing Opacity Parameters} refers to freezing geometry in Stage 2 to stabilize appearance optimization.
 }
    \vspace{-0.3cm}
    \footnotesize
    {
        \begin{tabular}{c|cc|ccc}
            \hline
            \multirow{2}{*}{\multirowcell{2}{\cellcolor{gray!20}\small Setting}} & 
            \multicolumn{2}{c|}{{\small Geometry}} & 
            \multicolumn{3}{c}{{\small Appearance}} \\
            & 
            \small CD$\downarrow$ & 
            \small F1$\uparrow$ & 
            \small PSNR$\uparrow$ & 
            \small SSIM$\uparrow$ & 
            \small LPIPS$\downarrow$ \\
            \hline
            \small Full Model & \small 1.85 & \small 0.95 & \small 39.08 & \small 0.989 & \small 0.020 \\
            \small w/o ASG & \small 1.86 & \small 0.96 & \small 37.54 & \small 0.987 & \small 0.023 \\
            \small w/o Normal Prior & \small 1.96 & \small 0.94 & \small 38.84 & \small 0.989 & \small 0.019 \\
            \small w/o De-light Prior & \small 1.87 & \small 0.95 & \small 38.95 & \small 0.989 & \small 0.020 \\
            \small w/o First-surface Depth & \small 1.89 & \small 0.95 & \small 39.08 & \small 0.989 & \small 0.020 \\
            \small w/o Fixing Opacity Params & \small 1.87 & \small 0.95 & \small 38.60 & \small 0.988 & \small 0.021 \\
            \hline
        \end{tabular}
    }
    \label{tab:ablation}
    \vspace{-0.45cm}
\end{table}

\noindent\textbf{Baseline Methods.}
We compared our method, TSGS, with current state-of-the-art neural surface reconstruction methods including NeuS~\cite{wang2021neus}, VolSDF~\cite{Yariv:2021:Volume}, and NeuralAngelo~\cite{li2023neuralangelo}. We also compared it with recently emerged reconstruction methods based on 3DGS, such as SuGaR~\cite{guedon2023sugar}, 2DGS~\cite{Huang2DGS2024}, GOF~\cite{yu2024gaussian}, and PGSR~\cite{chen2024pgsr}.
\subsection{Quantitative \& Qualitative Evaluation}
\label{subsec:evaluation}
As shown in Tab.~\ref{tab:geometric_quality}, Tab.~\ref{tab:image_quality}, and Tab.~\ref{tab:dtu_result}, our method achieves superior or highly competitive performance across different datasets.

\noindent\textbf{Geometric Accuracy.} TSGS demonstrates superior geometric accuracy on both transparent and opaque objects. On the challenging TransLab dataset (Tab.~\ref{tab:geometric_quality}), TSGS significantly improves geometric accuracy over the previous state-of-the-art (PGSR), achieving a \textbf{37.3\% reduction} in average chamfer distance (1.85 vs. 2.95) and an \textbf{8.0\% improvement} in average F1 score (0.95 vs. 0.88). This stems from our first-surface depth extraction and the two-stage training, which effectively resolve the transparency-depth dilemma. Furthermore, TSGS achieves state-of-the-art geometric accuracy on the standard opaque DTU dataset~\cite{jensen2014large} (Tab.~\ref{tab:dtu_result}), demonstrating the robustness and generalizability of our geometry learning approach.

\noindent\textbf{Novel View Synthesis.} Our method also yields high visual quality on TransLab (Tab.~\ref{tab:image_quality}). Compared to PGSR, TSGS improves average PSNR by 0.41dB (39.08 vs. 38.67) and slightly improves average SSIM (0.989 vs. 0.988). Our LPIPS (0.020) is comparable to the baseline (0.018). This suggests our staged approach achieves a good balance between geometric accuracy and visual quality.

\begin{figure}[t]
	\centering
	\includegraphics[width=\linewidth]{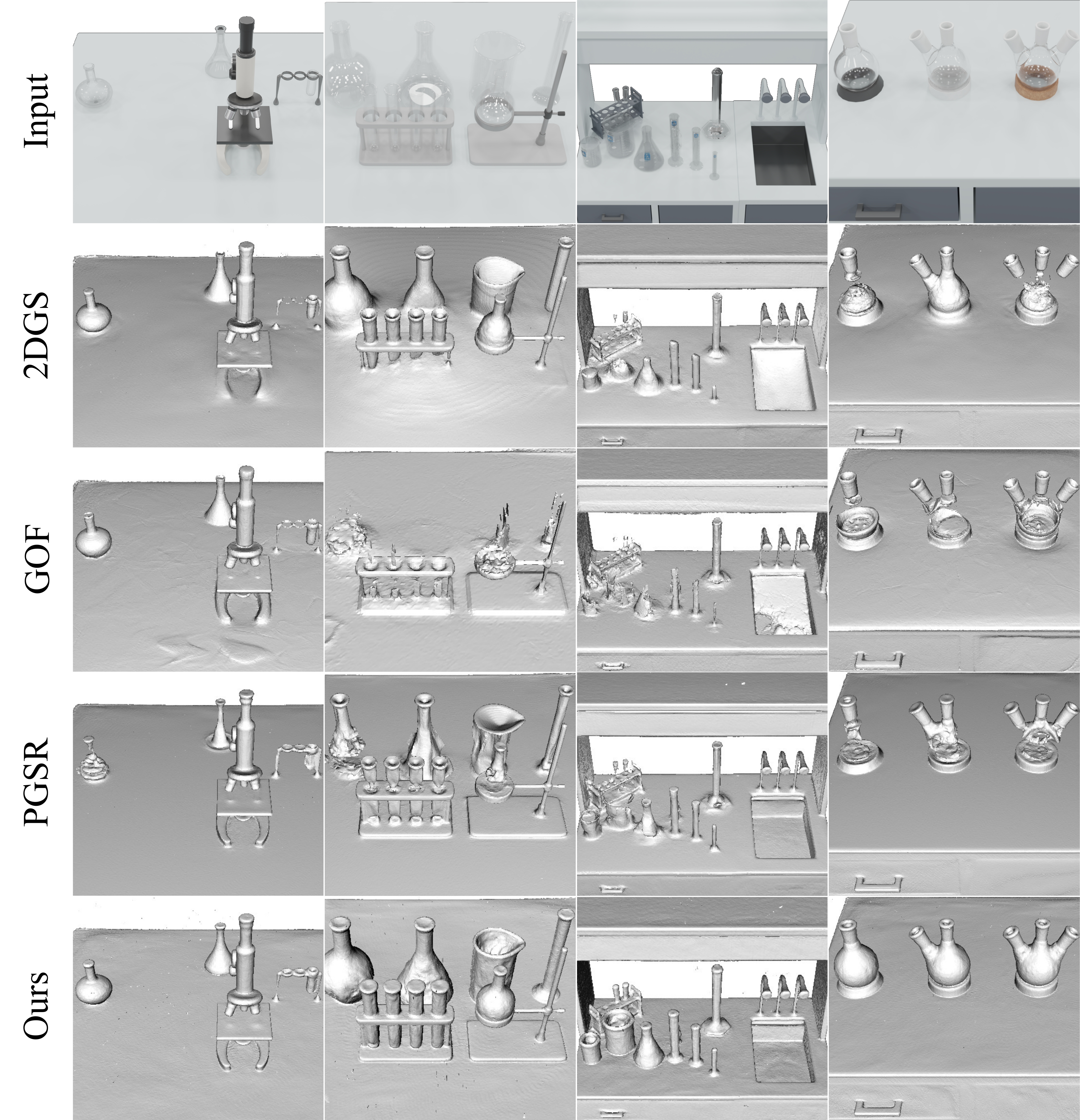}
    \vspace{-0.5cm}
	\caption{Comparison with other methods on the TransLab dataset. Our method accurately reconstructs transparent surfaces, while other methods fail to reconstruct the correct geometry, often resulting in holes or dips on transparent regions. (\S\ref{subsec:evaluation})}
    \vspace{-0.6cm}
	\label{fig:comparison_mesh}
\end{figure}
\noindent\textbf{Qualitative Results.}
Fig.~\ref{fig:comparison_mesh} demonstrates TSGS's capability in transparent object reconstruction. Compared to baselines, our method better preserves fine geometric details and achieves more realistic renderings. 
Due to the limited space, more qualitative results are shown in Appendix.
\subsection{Ablation Study}
\label{subsec:ablation-study}
We ablate key components of TSGS, and results are in Tab.~\ref{tab:ablation}.

\noindent\textbf{Normal Prior.} Excluding the Normal prior~\cite{ye2024stablenormal} from Stage 1 (\S\ref{subsec:stage1}) most significantly impacts geometry (CD: 1.85 vs. 1.96). This highlights the critical need for strong geometric priors to guide reconstruction for ill-posed transparent surfaces. 

\noindent\textbf{De-light Prior.} Removing the De-light prior~\cite{ye2024stablenormal} from Stage 1 (\S\ref{subsec:stage1}) slightly worsens geometry (CD: 1.85 vs. 1.87) and PSNR (39.08 vs. 38.95dB). The De-light prior helps handle strong highlights during initial geometry learning.

\noindent\textbf{First-surface Depth Extraction.} Replacing our first-surface depth extraction (\S\ref{subsec:first-surface-depth-extraction}) with unbiased depth extraction formulated in Eq.~\ref{eq:unbiased_depth} slightly increases CD (1.85 vs. 1.89). While rendered appearance metrics are unaffected, this step is vital for accurate first-surface geometry.

\noindent\textbf{Anisotropic Spherical Gaussians.} Replacing ASG (\S\ref{subsec:stage2}) with standard Gaussians significantly degrades appearance (PSNR: 39.08 vs. 37.54dB), demonstrating ASG's importance for modeling view-dependent specularities. Geometry is minimally affected as ASG primarily refines appearance in Stage 2.

\noindent\textbf{Fixing Opacity Parameters.} Optimizing all parameters jointly in Stage 2, instead of freezing opacity (\S\ref{subsec:stage2}), degrades both appearance (PSNR: 39.08 vs. 38.60dB) and geometry. This validates our two-stage strategy, where fixing opacity allows focused appearance refinement without compromising geometric accuracy.
% \vspace{-0.35cm}
\section{Conclusion}
\label{sec:conclusion}
Reconstructing transparent surfaces using 3D Gaussian Splatting is hampered by the transparency-depth dilemma, where appearance optimization conflicts with accurate geometry extraction. We introduced TSGS, a framework designed to resolve this dilemma. TSGS utilizes a two-stage training strategy to decouple geometry learning from appearance refinement. During inference, a novel first-surface depth extraction method robustly identifies the surface using a sliding window over rendering weights. Evaluated on our challenging new TransLab dataset, TSGS demonstrates state-of-the-art performance, significantly improving geometric accuracy (37.3\% chamfer distance reduction, 8.0\% F1 score improvement) over strong baselines while improving visual quality (+0.41dB PSNR). 
By effectively decoupling geometry and appearance and employing robust depth extraction, TSGS enables simultaneous high-fidelity reconstruction and rendering of transparent objects within the 3DGS framework.
\clearpage
\begin{acks}
This work was supported by the National Science and Technology Major Project (2023ZD0120803), National Natural Science Foundation of China (U2336212), Fundamental Research Funds for the Zhejiang Provincial Universities (226-2024-00208), "Pioneer" and "Leading Goose" R\&D Program of Zhejiang (No. 2025C02032), Earth System Big Data Platform of the School of Earth Sciences, Zhejiang University and Zhongguancun Academy, Beijing, China (20240313).
\end{acks}

\bibliographystyle{ACM-Reference-Format}
\bibliography{sample-base}


\begin{thebibliography}{84}

%%% the Web address.

\ifx \showCODEN    \undefined \def \showCODEN     #1{\unskip}     \fi
\ifx \showISBNx    \undefined \def \showISBNx     #1{\unskip}     \fi
\ifx \showISBNxiii \undefined \def \showISBNxiii  #1{\unskip}     \fi
\ifx \showISSN     \undefined \def \showISSN      #1{\unskip}     \fi
\ifx \showLCCN     \undefined \def \showLCCN      #1{\unskip}     \fi
\ifx \shownote     \undefined \def \shownote      #1{#1}          \fi
\ifx \showarticletitle \undefined \def \showarticletitle #1{#1}   \fi
\ifx \showURL      \undefined \def \showURL       {\relax}        \fi
% invisible to TeX
\providecommand\bibfield[2]{#2}
\providecommand\bibinfo[2]{#2}
\providecommand\natexlab[1]{#1}
\providecommand\showeprint[2][]{arXiv:#2}

\bibitem[Barnes et~al\mbox{.}(2009)]%
        {barnes2009patchmatch}
\bibfield{author}{\bibinfo{person}{Connelly Barnes}, \bibinfo{person}{Eli Shechtman}, \bibinfo{person}{Adam Finkelstein}, {and} \bibinfo{person}{Dan~B Goldman}.} \bibinfo{year}{2009}\natexlab{}.
\newblock \showarticletitle{PatchMatch: A randomized correspondence algorithm for structural image editing}.
\newblock \bibinfo{journal}{\emph{ACM Trans. Graph.}} \bibinfo{volume}{28}, \bibinfo{number}{3} (\bibinfo{year}{2009}), \bibinfo{pages}{24}.
\newblock


\bibitem[Barron et~al\mbox{.}(2021a)]%
        {barron2021mip}
\bibfield{author}{\bibinfo{person}{Jonathan~T Barron}, \bibinfo{person}{Ben Mildenhall}, \bibinfo{person}{Matthew Tancik}, \bibinfo{person}{Peter Hedman}, \bibinfo{person}{Ricardo Martin-Brualla}, {and} \bibinfo{person}{Pratul~P Srinivasan}.} \bibinfo{year}{2021}\natexlab{a}.
\newblock \showarticletitle{Mip-nerf: A multiscale representation for anti-aliasing neural radiance fields}. In \bibinfo{booktitle}{\emph{Proceedings of the IEEE/CVF International Conference on Computer Vision}}. \bibinfo{pages}{5855--5864}.
\newblock


\bibitem[Barron et~al\mbox{.}(2021b)]%
        {barron2021mipnerf}
\bibfield{author}{\bibinfo{person}{Jonathan~T. Barron}, \bibinfo{person}{Ben Mildenhall}, \bibinfo{person}{Matthew Tancik}, \bibinfo{person}{Peter Hedman}, \bibinfo{person}{Ricardo Martin-Brualla}, {and} \bibinfo{person}{Pratul~P. Srinivasan}.} \bibinfo{year}{2021}\natexlab{b}.
\newblock \showarticletitle{Mip-NeRF: A Multiscale Representation for Anti-Aliasing Neural Radiance Fields}.
\newblock \bibinfo{journal}{\emph{ICCV}} (\bibinfo{year}{2021}).
\newblock


\bibitem[Barron et~al\mbox{.}(2022)]%
        {barron2022mip}
\bibfield{author}{\bibinfo{person}{Jonathan~T Barron}, \bibinfo{person}{Ben Mildenhall}, \bibinfo{person}{Dor Verbin}, \bibinfo{person}{Pratul~P Srinivasan}, {and} \bibinfo{person}{Peter Hedman}.} \bibinfo{year}{2022}\natexlab{}.
\newblock \showarticletitle{Mip-nerf 360: Unbounded anti-aliased neural radiance fields}. In \bibinfo{booktitle}{\emph{Proceedings of the IEEE/CVF conference on computer vision and pattern recognition}}. \bibinfo{pages}{5470--5479}.
\newblock


\bibitem[Barron et~al\mbox{.}(2023)]%
        {barron2023zip}
\bibfield{author}{\bibinfo{person}{Jonathan~T Barron}, \bibinfo{person}{Ben Mildenhall}, \bibinfo{person}{Dor Verbin}, \bibinfo{person}{Pratul~P Srinivasan}, {and} \bibinfo{person}{Peter Hedman}.} \bibinfo{year}{2023}\natexlab{}.
\newblock \showarticletitle{Zip-nerf: Anti-aliased grid-based neural radiance fields}. In \bibinfo{booktitle}{\emph{Proceedings of the IEEE/CVF International Conference on Computer Vision}}. \bibinfo{pages}{19697--19705}.
\newblock


\bibitem[Boissonnat(1984)]%
        {boissonnat1984geometric}
\bibfield{author}{\bibinfo{person}{Jean-Daniel Boissonnat}.} \bibinfo{year}{1984}\natexlab{}.
\newblock \showarticletitle{Geometric structures for three-dimensional shape representation}.
\newblock \bibinfo{journal}{\emph{ACM Trans. Graph.}} \bibinfo{volume}{3}, \bibinfo{number}{4} (\bibinfo{date}{Oct.} \bibinfo{year}{1984}), \bibinfo{pages}{266–286}.
\newblock
\showISSN{0730-0301}
\href{https://doi.org/10.1145/357346.357349}{doi:\nolinkurl{10.1145/357346.357349}}


\bibitem[Campos et~al\mbox{.}(2021)]%
        {campos2021orb}
\bibfield{author}{\bibinfo{person}{Carlos Campos}, \bibinfo{person}{Richard Elvira}, \bibinfo{person}{Juan J~G{\'o}mez Rodr{\'\i}guez}, \bibinfo{person}{Jos{\'e}~MM Montiel}, {and} \bibinfo{person}{Juan~D Tard{\'o}s}.} \bibinfo{year}{2021}\natexlab{}.
\newblock \showarticletitle{Orb-slam3: An accurate open-source library for visual, visual--inertial, and multimap slam}.
\newblock \bibinfo{journal}{\emph{IEEE Transactions on Robotics}} \bibinfo{volume}{37}, \bibinfo{number}{6} (\bibinfo{year}{2021}), \bibinfo{pages}{1874--1890}.
\newblock


\bibitem[Cazals and Giesen(2006)]%
        {cazals2006delaunay}
\bibfield{author}{\bibinfo{person}{Fr{\'e}d{\'e}ric Cazals} {and} \bibinfo{person}{Joachim Giesen}.} \bibinfo{year}{2006}\natexlab{}.
\newblock \showarticletitle{Delaunay triangulation based surface reconstruction}.
\newblock In \bibinfo{booktitle}{\emph{Effective computational geometry for curves and surfaces}}. \bibinfo{publisher}{Springer}, \bibinfo{pages}{231--276}.
\newblock


\bibitem[Chen et~al\mbox{.}(2024b)]%
        {chen2024pgsr}
\bibfield{author}{\bibinfo{person}{Danpeng Chen}, \bibinfo{person}{Hai Li}, \bibinfo{person}{Weicai Ye}, \bibinfo{person}{Yifan Wang}, \bibinfo{person}{Weijian Xie}, \bibinfo{person}{Shangjin Zhai}, \bibinfo{person}{Nan Wang}, \bibinfo{person}{Haomin Liu}, \bibinfo{person}{Hujun Bao}, {and} \bibinfo{person}{Guofeng Zhang}.} \bibinfo{year}{2024}\natexlab{b}.
\newblock \showarticletitle{PGSR: Planar-based Gaussian Splatting for Efficient and High-Fidelity Surface Reconstruction}.
\newblock \bibinfo{journal}{\emph{IEEE Transactions on Visualization and Computer Graphics}}  \bibinfo{volume}{PP} (\bibinfo{date}{01} \bibinfo{year}{2024}), \bibinfo{pages}{1--12}.
\newblock
\href{https://doi.org/10.1109/TVCG.2024.3494046}{doi:\nolinkurl{10.1109/TVCG.2024.3494046}}


\bibitem[Chen et~al\mbox{.}(2021)]%
        {chen2021rnin}
\bibfield{author}{\bibinfo{person}{Danpeng Chen}, \bibinfo{person}{Nan Wang}, \bibinfo{person}{Runsen Xu}, \bibinfo{person}{Weijian Xie}, \bibinfo{person}{Hujun Bao}, {and} \bibinfo{person}{Guofeng Zhang}.} \bibinfo{year}{2021}\natexlab{}.
\newblock \showarticletitle{Rnin-vio: Robust neural inertial navigation aided visual-inertial odometry in challenging scenes}. In \bibinfo{booktitle}{\emph{2021 IEEE International Symposium on Mixed and Augmented Reality (ISMAR)}}. IEEE, \bibinfo{pages}{275--283}.
\newblock


\bibitem[Chen et~al\mbox{.}(2022a)]%
        {chen2022vip}
\bibfield{author}{\bibinfo{person}{Danpeng Chen}, \bibinfo{person}{Shuai Wang}, \bibinfo{person}{Weijian Xie}, \bibinfo{person}{Shangjin Zhai}, \bibinfo{person}{Nan Wang}, \bibinfo{person}{Hujun Bao}, {and} \bibinfo{person}{Guofeng Zhang}.} \bibinfo{year}{2022}\natexlab{a}.
\newblock \showarticletitle{Vip-slam: An efficient tightly-coupled rgb-d visual inertial planar slam}. In \bibinfo{booktitle}{\emph{2022 International Conference on Robotics and Automation (ICRA)}}. IEEE, \bibinfo{pages}{5615--5621}.
\newblock


\bibitem[Chen et~al\mbox{.}(2022b)]%
        {chen2022clearpose}
\bibfield{author}{\bibinfo{person}{Xiaotong Chen}, \bibinfo{person}{Huijie Zhang}, \bibinfo{person}{Zeren Yu}, \bibinfo{person}{Anthony Opipari}, {and} \bibinfo{person}{Odest Chadwicke~Jenkins}.} \bibinfo{year}{2022}\natexlab{b}.
\newblock \showarticletitle{ClearPose: Large-scale Transparent Object Dataset and;Benchmark}. In \bibinfo{booktitle}{\emph{Computer Vision - ECCV 2022: 17th European Conference, Tel Aviv, Israel, October 23-27, 2022, Proceedings, Part VIII}} (Tel Aviv, Israel). \bibinfo{publisher}{Springer-Verlag}, \bibinfo{address}{Berlin, Heidelberg}, \bibinfo{pages}{381--396}.
\newblock
\showISBNx{978-3-031-20073-1}
\href{https://doi.org/10.1007/978-3-031-20074-8_22}{doi:\nolinkurl{10.1007/978-3-031-20074-8_22}}


\bibitem[Chen et~al\mbox{.}(2024a)]%
        {chen2024meshanything}
\bibfield{author}{\bibinfo{person}{Yiwen Chen}, \bibinfo{person}{Tong He}, \bibinfo{person}{Di Huang}, \bibinfo{person}{Weicai Ye}, \bibinfo{person}{Sijin Chen}, \bibinfo{person}{Jiaxiang Tang}, \bibinfo{person}{Xin Chen}, \bibinfo{person}{Zhongang Cai}, \bibinfo{person}{Lei Yang}, \bibinfo{person}{Gang Yu}, {et~al\mbox{.}}} \bibinfo{year}{2024}\natexlab{a}.
\newblock \showarticletitle{{MeshAnything: Artist-Created Mesh Generation with Autoregressive Transformers}}.
\newblock \bibinfo{journal}{\emph{arXiv preprint arXiv:2406.10163}} (\bibinfo{year}{2024}).
\newblock


\bibitem[Cheng et~al\mbox{.}(2024)]%
        {gaussianpro}
\bibfield{author}{\bibinfo{person}{Kai Cheng}, \bibinfo{person}{Xiaoxiao Long}, \bibinfo{person}{Kaizhi Yang}, \bibinfo{person}{Yao Yao}, \bibinfo{person}{Wei Yin}, \bibinfo{person}{Yuexin Ma}, \bibinfo{person}{Wenping Wang}, {and} \bibinfo{person}{Xuejin Chen}.} \bibinfo{year}{2024}\natexlab{}.
\newblock \showarticletitle{GaussianPro: 3D Gaussian splatting with progressive propagation}. In \bibinfo{booktitle}{\emph{Proceedings of the 41st International Conference on Machine Learning}} (Vienna, Austria) \emph{(\bibinfo{series}{ICML'24})}. \bibinfo{publisher}{JMLR.org}, Article \bibinfo{articleno}{320}, \bibinfo{numpages}{18}~pages.
\newblock


\bibitem[Choy et~al\mbox{.}(2016)]%
        {Choy:2016:3d}
\bibfield{author}{\bibinfo{person}{Christopher~B. Choy}, \bibinfo{person}{Danfei Xu}, \bibinfo{person}{JunYoung Gwak}, \bibinfo{person}{Kevin Chen}, {and} \bibinfo{person}{Silvio Savarese}.} \bibinfo{year}{2016}\natexlab{}.
\newblock \showarticletitle{{3D-R2N2}: {A} Unified Approach for Single and Multi-view {3D} Object Reconstruction}. In \bibinfo{booktitle}{\emph{European Conference on Computer Vision}}, Vol.~\bibinfo{volume}{9912}. \bibinfo{pages}{628--644}.
\newblock


\bibitem[Curless and Levoy(1996)]%
        {curless1996volumetric}
\bibfield{author}{\bibinfo{person}{Brian Curless} {and} \bibinfo{person}{Marc Levoy}.} \bibinfo{year}{1996}\natexlab{}.
\newblock \showarticletitle{A volumetric method for building complex models from range images}. In \bibinfo{booktitle}{\emph{SIGGRAPH}}. \bibinfo{pages}{303--312}.
\newblock


\bibitem[Deng et~al\mbox{.}(2022)]%
        {deng2022fov}
\bibfield{author}{\bibinfo{person}{Nianchen Deng}, \bibinfo{person}{Zhenyi He}, \bibinfo{person}{Jiannan Ye}, \bibinfo{person}{Budmonde Duinkharjav}, \bibinfo{person}{Praneeth Chakravarthula}, \bibinfo{person}{Xubo Yang}, {and} \bibinfo{person}{Qi Sun}.} \bibinfo{year}{2022}\natexlab{}.
\newblock \showarticletitle{Fov-nerf: Foveated neural radiance fields for virtual reality}.
\newblock \bibinfo{journal}{\emph{IEEE Transactions on Visualization and Computer Graphics}} \bibinfo{volume}{28}, \bibinfo{number}{11} (\bibinfo{year}{2022}), \bibinfo{pages}{3854--3864}.
\newblock


\bibitem[Deng et~al\mbox{.}(2024)]%
        {Deng:CVPR2024}
\bibfield{author}{\bibinfo{person}{Weijian Deng}, \bibinfo{person}{Dylan Campbell}, \bibinfo{person}{Chunyi Sun}, \bibinfo{person}{Shubham Kanitkar}, \bibinfo{person}{Matthew Shaffer}, {and} \bibinfo{person}{Stephen Gould}.} \bibinfo{year}{2024}\natexlab{}.
\newblock \showarticletitle{Differentiable Neural Surface Refinement for Transparent Objects}. In \bibinfo{booktitle}{\emph{CVPR}}.
\newblock


\bibitem[Edelsbrunner and M\"{u}cke(1994)]%
        {edelsbrunner1994three}
\bibfield{author}{\bibinfo{person}{Herbert Edelsbrunner} {and} \bibinfo{person}{Ernst~P. M\"{u}cke}.} \bibinfo{year}{1994}\natexlab{}.
\newblock \showarticletitle{Three-dimensional alpha shapes}.
\newblock \bibinfo{journal}{\emph{ACM Trans. Graph.}} \bibinfo{volume}{13}, \bibinfo{number}{1} (\bibinfo{date}{Jan.} \bibinfo{year}{1994}), \bibinfo{pages}{43–72}.
\newblock
\showISSN{0730-0301}
\href{https://doi.org/10.1145/174462.156635}{doi:\nolinkurl{10.1145/174462.156635}}


\bibitem[Fan et~al\mbox{.}(2017)]%
        {Fan:2017:Point}
\bibfield{author}{\bibinfo{person}{Haoqiang Fan}, \bibinfo{person}{Hao Su}, {and} \bibinfo{person}{Leonidas~J. Guibas}.} \bibinfo{year}{2017}\natexlab{}.
\newblock \showarticletitle{A Point Set Generation Network for {3D} Object Reconstruction from a Single Image}. In \bibinfo{booktitle}{\emph{{IEEE} Conference on Computer Vision and Pattern Recognition}}. \bibinfo{pages}{2463--2471}.
\newblock


\bibitem[Furukawa et~al\mbox{.}(2015)]%
        {furukawa2015multi}
\bibfield{author}{\bibinfo{person}{Yasutaka Furukawa}, \bibinfo{person}{Carlos Hern{\'a}ndez}, {et~al\mbox{.}}} \bibinfo{year}{2015}\natexlab{}.
\newblock \showarticletitle{Multi-view stereo: A tutorial}.
\newblock \bibinfo{journal}{\emph{Foundations and trends{\textregistered} in Computer Graphics and Vision}} \bibinfo{volume}{9}, \bibinfo{number}{1-2} (\bibinfo{year}{2015}), \bibinfo{pages}{1--148}.
\newblock


\bibitem[Gao et~al\mbox{.}(2023)]%
        {gao2023transparent}
\bibfield{author}{\bibinfo{person}{Fangzhou Gao}, \bibinfo{person}{Lianghao Zhang}, \bibinfo{person}{Li Wang}, \bibinfo{person}{Jiamin Cheng}, {and} \bibinfo{person}{Jiawan Zhang}.} \bibinfo{year}{2023}\natexlab{}.
\newblock \showarticletitle{Transparent Object Reconstruction via Implicit Differentiable Refraction Rendering}. In \bibinfo{booktitle}{\emph{SIGGRAPH Asia 2023 Conference Papers}} (Sydney, NSW, Australia) \emph{(\bibinfo{series}{SA '23})}. \bibinfo{publisher}{Association for Computing Machinery}, \bibinfo{address}{New York, NY, USA}, Article \bibinfo{articleno}{57}, \bibinfo{numpages}{11}~pages.
\newblock
\showISBNx{9798400703157}
\href{https://doi.org/10.1145/3610548.3618236}{doi:\nolinkurl{10.1145/3610548.3618236}}


\bibitem[Gao et~al\mbox{.}(2024)]%
        {gao2024luminat2x}
\bibfield{author}{\bibinfo{person}{Peng Gao}, \bibinfo{person}{Le Zhuo}, \bibinfo{person}{Dongyang Liu}, \bibinfo{person}{Ruoyi Du}, \bibinfo{person}{Xu Luo}, \bibinfo{person}{Longtian Qiu}, \bibinfo{person}{Yuhang Zhang}, \bibinfo{person}{Chen Lin}, \bibinfo{person}{Rongjie Huang}, \bibinfo{person}{Shijie Geng}, \bibinfo{person}{Renrui Zhang}, \bibinfo{person}{Junlin Xi}, \bibinfo{person}{Wenqi Shao}, \bibinfo{person}{Zhengkai Jiang}, \bibinfo{person}{Tianshuo Yang}, \bibinfo{person}{Weicai Ye}, \bibinfo{person}{He Tong}, \bibinfo{person}{Jingwen He}, \bibinfo{person}{Yu Qiao}, {and} \bibinfo{person}{Hongsheng Li}.} \bibinfo{year}{2024}\natexlab{}.
\newblock \showarticletitle{Lumina-T2X: Transforming Text into Any Modality, Resolution, and Duration via Flow-based Large Diffusion Transformers}.
\newblock \bibinfo{journal}{\emph{arXiv preprint arxiv:2405.05945}} (\bibinfo{year}{2024}).
\newblock


\bibitem[Gu{\'e}don and Lepetit(2024)]%
        {guedon2023sugar}
\bibfield{author}{\bibinfo{person}{Antoine Gu{\'e}don} {and} \bibinfo{person}{Vincent Lepetit}.} \bibinfo{year}{2024}\natexlab{}.
\newblock \showarticletitle{SuGaR: Surface-Aligned Gaussian Splatting for Efficient 3D Mesh Reconstruction and High-Quality Mesh Rendering}.
\newblock \bibinfo{journal}{\emph{CVPR}} (\bibinfo{year}{2024}).
\newblock


\bibitem[Huang et~al\mbox{.}(2024b)]%
        {Huang2DGS2024}
\bibfield{author}{\bibinfo{person}{Binbin Huang}, \bibinfo{person}{Zehao Yu}, \bibinfo{person}{Anpei Chen}, \bibinfo{person}{Andreas Geiger}, {and} \bibinfo{person}{Shenghua Gao}.} \bibinfo{year}{2024}\natexlab{b}.
\newblock \showarticletitle{2D Gaussian Splatting for Geometrically Accurate Radiance Fields}. In \bibinfo{booktitle}{\emph{SIGGRAPH 2024 Conference Papers}}. \bibinfo{publisher}{Association for Computing Machinery}.
\newblock
\href{https://doi.org/10.1145/3641519.3657428}{doi:\nolinkurl{10.1145/3641519.3657428}}


\bibitem[Huang et~al\mbox{.}(2024a)]%
        {huang2024nerfdet++}
\bibfield{author}{\bibinfo{person}{Chenxi Huang}, \bibinfo{person}{Yuenan Hou}, \bibinfo{person}{Weicai Ye}, \bibinfo{person}{Di Huang}, \bibinfo{person}{Xiaoshui Huang}, \bibinfo{person}{Binbin Lin}, \bibinfo{person}{Deng Cai}, {and} \bibinfo{person}{Wanli Ouyang}.} \bibinfo{year}{2024}\natexlab{a}.
\newblock \showarticletitle{NeRF-Det++: Incorporating Semantic Cues and Perspective-aware Depth Supervision for Indoor Multi-View 3D Detection}.
\newblock \bibinfo{journal}{\emph{arXiv preprint arXiv:2402.14464}} (\bibinfo{year}{2024}).
\newblock


\bibitem[Jensen et~al\mbox{.}(2014)]%
        {jensen2014large}
\bibfield{author}{\bibinfo{person}{Rasmus Jensen}, \bibinfo{person}{Anders Dahl}, \bibinfo{person}{George Vogiatzis}, \bibinfo{person}{Engin Tola}, {and} \bibinfo{person}{Henrik Aan{\ae}s}.} \bibinfo{year}{2014}\natexlab{}.
\newblock \showarticletitle{Large scale multi-view stereopsis evaluation}. In \bibinfo{booktitle}{\emph{Proceedings of the IEEE conference on computer vision and pattern recognition}}. \bibinfo{pages}{406--413}.
\newblock


\bibitem[Jiang et~al\mbox{.}(2024a)]%
        {jiang2024trex2}
\bibfield{author}{\bibinfo{person}{Qing Jiang}, \bibinfo{person}{Feng Li}, \bibinfo{person}{Zhaoyang Zeng}, \bibinfo{person}{Tianhe Ren}, \bibinfo{person}{Shilong Liu}, {and} \bibinfo{person}{Lei Zhang}.} \bibinfo{year}{2024}\natexlab{a}.
\newblock \bibinfo{title}{T-Rex2: Towards Generic Object Detection via Text-Visual Prompt Synergy}.
\newblock
\showeprint[arxiv]{2403.14610}~[cs.CV]


\bibitem[Jiang et~al\mbox{.}(2024b)]%
        {gaussianshader}
\bibfield{author}{\bibinfo{person}{Yingwenqi Jiang}, \bibinfo{person}{Jiadong Tu}, \bibinfo{person}{Yuan Liu}, \bibinfo{person}{Xifeng Gao}, \bibinfo{person}{Xiaoxiao Long}, \bibinfo{person}{Wenping Wang}, {and} \bibinfo{person}{Yuexin Ma}.} \bibinfo{year}{2024}\natexlab{b}.
\newblock \showarticletitle{GaussianShader: 3D Gaussian Splatting with Shading Functions for Reflective Surfaces}. In \bibinfo{booktitle}{\emph{2024 IEEE/CVF Conference on Computer Vision and Pattern Recognition (CVPR)}}. \bibinfo{pages}{5322--5332}.
\newblock
\href{https://doi.org/10.1109/CVPR52733.2024.00509}{doi:\nolinkurl{10.1109/CVPR52733.2024.00509}}


\bibitem[Kazhdan et~al\mbox{.}(2006)]%
        {kazhdan2006poisson}
\bibfield{author}{\bibinfo{person}{Michael Kazhdan}, \bibinfo{person}{Matthew Bolitho}, {and} \bibinfo{person}{Hugues Hoppe}.} \bibinfo{year}{2006}\natexlab{}.
\newblock \showarticletitle{Poisson surface reconstruction}. In \bibinfo{booktitle}{\emph{Proceedings of the Fourth Eurographics Symposium on Geometry Processing}} (Cagliari, Sardinia, Italy) \emph{(\bibinfo{series}{SGP '06})}. \bibinfo{publisher}{Eurographics Association}, \bibinfo{address}{Goslar, DEU}, \bibinfo{pages}{61–70}.
\newblock
\showISBNx{3905673363}


\bibitem[Kerbl et~al\mbox{.}(2023)]%
        {kerbl20233d}
\bibfield{author}{\bibinfo{person}{Bernhard Kerbl}, \bibinfo{person}{Georgios Kopanas}, \bibinfo{person}{Thomas Leimk{\"u}hler}, {and} \bibinfo{person}{George Drettakis}.} \bibinfo{year}{2023}\natexlab{}.
\newblock \showarticletitle{3D Gaussian Splatting for Real-Time Radiance Field Rendering}. In \bibinfo{booktitle}{\emph{ACM Transactions on Graphics}}, Vol.~\bibinfo{volume}{42}.
\newblock


\bibitem[Kim et~al\mbox{.}(2025)]%
        {kim2025transplatsurfaceembeddingguided3d}
\bibfield{author}{\bibinfo{person}{Jeongyun Kim}, \bibinfo{person}{Jeongho Noh}, \bibinfo{person}{Dong-Guw Lee}, {and} \bibinfo{person}{Ayoung Kim}.} \bibinfo{year}{2025}\natexlab{}.
\newblock \bibinfo{title}{TranSplat: Surface Embedding-guided 3D Gaussian Splatting for Transparent Object Manipulation}.
\newblock
\showeprint[arxiv]{2502.07840}~[cs.CV]
\urldef\tempurl%
\url{https://arxiv.org/abs/2502.07840}
\showURL{%
\tempurl}


\bibitem[Kirillov et~al\mbox{.}(2023)]%
        {kirillov2023segany}
\bibfield{author}{\bibinfo{person}{Alexander Kirillov}, \bibinfo{person}{Eric Mintun}, \bibinfo{person}{Nikhila Ravi}, \bibinfo{person}{Hanzi Mao}, \bibinfo{person}{Chloe Rolland}, \bibinfo{person}{Laura Gustafson}, \bibinfo{person}{Tete Xiao}, \bibinfo{person}{Spencer Whitehead}, \bibinfo{person}{Alexander~C. Berg}, \bibinfo{person}{Wan-Yen Lo}, \bibinfo{person}{Piotr Doll{\'a}r}, {and} \bibinfo{person}{Ross Girshick}.} \bibinfo{year}{2023}\natexlab{}.
\newblock \showarticletitle{Segment Anything}.
\newblock \bibinfo{journal}{\emph{arXiv:2304.02643}} (\bibinfo{year}{2023}).
\newblock


\bibitem[Knapitsch et~al\mbox{.}(2017)]%
        {knapitsch2017tanks}
\bibfield{author}{\bibinfo{person}{Arno Knapitsch}, \bibinfo{person}{Jaesik Park}, \bibinfo{person}{Qian-Yi Zhou}, {and} \bibinfo{person}{Vladlen Koltun}.} \bibinfo{year}{2017}\natexlab{}.
\newblock \showarticletitle{Tanks and temples: Benchmarking large-scale scene reconstruction}.
\newblock \bibinfo{journal}{\emph{ACM Transactions on Graphics (ToG)}} \bibinfo{volume}{36}, \bibinfo{number}{4} (\bibinfo{year}{2017}), \bibinfo{pages}{1--13}.
\newblock


\bibitem[Kutulakos and Seitz(2000)]%
        {kutulakos2000theory}
\bibfield{author}{\bibinfo{person}{Kiriakos~N Kutulakos} {and} \bibinfo{person}{Steven~M Seitz}.} \bibinfo{year}{2000}\natexlab{}.
\newblock \showarticletitle{A theory of shape by space carving}.
\newblock \bibinfo{journal}{\emph{International journal of computer vision}}  \bibinfo{volume}{38} (\bibinfo{year}{2000}), \bibinfo{pages}{199--218}.
\newblock


\bibitem[Levin(2004)]%
        {levin2004mesh}
\bibfield{author}{\bibinfo{person}{David Levin}.} \bibinfo{year}{2004}\natexlab{}.
\newblock \showarticletitle{Mesh-independent surface interpolation}. In \bibinfo{booktitle}{\emph{Geometric modeling for scientific visualization}}. Springer, \bibinfo{pages}{37--49}.
\newblock


\bibitem[Lhuillier and Quan(2005)]%
        {lhuillier2005quasi}
\bibfield{author}{\bibinfo{person}{Maxime Lhuillier} {and} \bibinfo{person}{Long Quan}.} \bibinfo{year}{2005}\natexlab{}.
\newblock \showarticletitle{A quasi-dense approach to surface reconstruction from uncalibrated images}.
\newblock \bibinfo{journal}{\emph{IEEE transactions on pattern analysis and machine intelligence}} \bibinfo{volume}{27}, \bibinfo{number}{3} (\bibinfo{year}{2005}), \bibinfo{pages}{418--433}.
\newblock


\bibitem[Li et~al\mbox{.}(2025)]%
        {li2025cargsaddressingreflectivetransparent}
\bibfield{author}{\bibinfo{person}{Congcong Li}, \bibinfo{person}{Jin Wang}, \bibinfo{person}{Xiaomeng Wang}, \bibinfo{person}{Xingchen Zhou}, \bibinfo{person}{Wei Wu}, \bibinfo{person}{Yuzhi Zhang}, {and} \bibinfo{person}{Tongyi Cao}.} \bibinfo{year}{2025}\natexlab{}.
\newblock \bibinfo{title}{Car-GS: Addressing Reflective and Transparent Surface Challenges in 3D Car Reconstruction}.
\newblock
\showeprint[arxiv]{2501.11020}~[cs.CV]
\urldef\tempurl%
\url{https://arxiv.org/abs/2501.11020}
\showURL{%
\tempurl}


\bibitem[Li et~al\mbox{.}(2022)]%
        {li2022vox}
\bibfield{author}{\bibinfo{person}{Hai Li}, \bibinfo{person}{Xingrui Yang}, \bibinfo{person}{Hongjia Zhai}, \bibinfo{person}{Yuqian Liu}, \bibinfo{person}{Hujun Bao}, {and} \bibinfo{person}{Guofeng Zhang}.} \bibinfo{year}{2022}\natexlab{}.
\newblock \showarticletitle{Vox-surf: Voxel-based implicit surface representation}.
\newblock \bibinfo{journal}{\emph{IEEE Transactions on Visualization and Computer Graphics}} (\bibinfo{year}{2022}).
\newblock


\bibitem[Li et~al\mbox{.}(2020)]%
        {li2020saliency}
\bibfield{author}{\bibinfo{person}{Hai Li}, \bibinfo{person}{Weicai Ye}, \bibinfo{person}{Guofeng Zhang}, \bibinfo{person}{Sanyuan Zhang}, {and} \bibinfo{person}{Hujun Bao}.} \bibinfo{year}{2020}\natexlab{}.
\newblock \showarticletitle{Saliency guided subdivision for single-view mesh reconstruction}. In \bibinfo{booktitle}{\emph{2020 International Conference on 3D Vision (3DV)}}. IEEE, \bibinfo{pages}{1098--1107}.
\newblock


\bibitem[Li et~al\mbox{.}(2023)]%
        {li2023neuralangelo}
\bibfield{author}{\bibinfo{person}{Zhaoshuo Li}, \bibinfo{person}{Thomas M{\"u}ller}, \bibinfo{person}{Alex Evans}, \bibinfo{person}{Russell~H Taylor}, \bibinfo{person}{Mathias Unberath}, \bibinfo{person}{Ming-Yu Liu}, {and} \bibinfo{person}{Chen-Hsuan Lin}.} \bibinfo{year}{2023}\natexlab{}.
\newblock \showarticletitle{Neuralangelo: High-fidelity neural surface reconstruction}. In \bibinfo{booktitle}{\emph{Proceedings of the IEEE/CVF Conference on Computer Vision and Pattern Recognition}}. \bibinfo{pages}{8456--8465}.
\newblock


\bibitem[Lin et~al\mbox{.}(2018)]%
        {Lin:2018:Learning}
\bibfield{author}{\bibinfo{person}{Chen{-}Hsuan Lin}, \bibinfo{person}{Chen Kong}, {and} \bibinfo{person}{Simon Lucey}.} \bibinfo{year}{2018}\natexlab{}.
\newblock \showarticletitle{Learning Efficient Point Cloud Generation for Dense {3D} Object Reconstruction}. In \bibinfo{booktitle}{\emph{Conference on Artificial Intelligence}}. \bibinfo{pages}{7114--7121}.
\newblock


\bibitem[Liu et~al\mbox{.}(2020)]%
        {Liu:2020:Neural}
\bibfield{author}{\bibinfo{person}{Lingjie Liu}, \bibinfo{person}{Jiatao Gu}, \bibinfo{person}{Kyaw~Zaw Lin}, \bibinfo{person}{Tat{-}Seng Chua}, {and} \bibinfo{person}{Christian Theobalt}.} \bibinfo{year}{2020}\natexlab{}.
\newblock \showarticletitle{Neural Sparse Voxel Fields}. In \bibinfo{booktitle}{\emph{Advances in Neural Information Processing Systems}}. \bibinfo{pages}{15651--15663}.
\newblock


\bibitem[Liu et~al\mbox{.}(2023)]%
        {liu2023grounding}
\bibfield{author}{\bibinfo{person}{Shilong Liu}, \bibinfo{person}{Zhaoyang Zeng}, \bibinfo{person}{Tianhe Ren}, \bibinfo{person}{Feng Li}, \bibinfo{person}{Hao Zhang}, \bibinfo{person}{Jie Yang}, \bibinfo{person}{Chunyuan Li}, \bibinfo{person}{Jianwei Yang}, \bibinfo{person}{Hang Su}, \bibinfo{person}{Jun Zhu}, {et~al\mbox{.}}} \bibinfo{year}{2023}\natexlab{}.
\newblock \showarticletitle{Grounding dino: Marrying dino with grounded pre-training for open-set object detection}.
\newblock \bibinfo{journal}{\emph{arXiv preprint arXiv:2303.05499}} (\bibinfo{year}{2023}).
\newblock


\bibitem[Lorensen and Cline(1987)]%
        {lorensen1987marching}
\bibfield{author}{\bibinfo{person}{William~E. Lorensen} {and} \bibinfo{person}{Harvey~E. Cline}.} \bibinfo{year}{1987}\natexlab{}.
\newblock \showarticletitle{Marching cubes: A high resolution 3D surface construction algorithm}.
\newblock \bibinfo{journal}{\emph{SIGGRAPH Comput. Graph.}} \bibinfo{volume}{21}, \bibinfo{number}{4} (\bibinfo{date}{Aug.} \bibinfo{year}{1987}), \bibinfo{pages}{163–169}.
\newblock
\showISSN{0097-8930}
\href{https://doi.org/10.1145/37402.37422}{doi:\nolinkurl{10.1145/37402.37422}}


\bibitem[Lyu et~al\mbox{.}(2020)]%
        {lyu2020differentiablerefraction-tracing}
\bibfield{author}{\bibinfo{person}{Jiahui Lyu}, \bibinfo{person}{Bojian Wu}, \bibinfo{person}{Dani Lischinski}, \bibinfo{person}{Daniel Cohen-Or}, {and} \bibinfo{person}{Hui Huang}.} \bibinfo{year}{2020}\natexlab{}.
\newblock \showarticletitle{Differentiable refraction-tracing for mesh reconstruction of transparent objects}.
\newblock \bibinfo{journal}{\emph{ACM Trans. Graph.}} \bibinfo{volume}{39}, \bibinfo{number}{6}, Article \bibinfo{articleno}{195} (\bibinfo{date}{Nov.} \bibinfo{year}{2020}), \bibinfo{numpages}{13}~pages.
\newblock
\showISSN{0730-0301}
\href{https://doi.org/10.1145/3414685.3417815}{doi:\nolinkurl{10.1145/3414685.3417815}}


\bibitem[Mescheder et~al\mbox{.}(2019)]%
        {Mescheder:2019:Occupancy}
\bibfield{author}{\bibinfo{person}{Lars~M. Mescheder}, \bibinfo{person}{Michael Oechsle}, \bibinfo{person}{Michael Niemeyer}, \bibinfo{person}{Sebastian Nowozin}, {and} \bibinfo{person}{Andreas Geiger}.} \bibinfo{year}{2019}\natexlab{}.
\newblock \showarticletitle{Occupancy Networks: Learning {3D} Reconstruction in Function Space}. In \bibinfo{booktitle}{\emph{{IEEE} Conference on Computer Vision and Pattern Recognition}}. \bibinfo{pages}{4460--4470}.
\newblock


\bibitem[Mildenhall et~al\mbox{.}(2021)]%
        {mildenhall2021nerf}
\bibfield{author}{\bibinfo{person}{Ben Mildenhall}, \bibinfo{person}{Pratul~P Srinivasan}, \bibinfo{person}{Matthew Tancik}, \bibinfo{person}{Jonathan~T Barron}, \bibinfo{person}{Ravi Ramamoorthi}, {and} \bibinfo{person}{Ren Ng}.} \bibinfo{year}{2021}\natexlab{}.
\newblock \showarticletitle{Nerf: Representing scenes as neural radiance fields for view synthesis}.
\newblock \bibinfo{journal}{\emph{Commun. ACM}} \bibinfo{volume}{65}, \bibinfo{number}{1} (\bibinfo{year}{2021}), \bibinfo{pages}{99--106}.
\newblock


\bibitem[Ming et~al\mbox{.}(2022)]%
        {ming2022idf}
\bibfield{author}{\bibinfo{person}{Yuhang Ming}, \bibinfo{person}{Weicai Ye}, {and} \bibinfo{person}{Andrew Calway}.} \bibinfo{year}{2022}\natexlab{}.
\newblock \showarticletitle{idf-slam: End-to-end rgb-d slam with neural implicit mapping and deep feature tracking}.
\newblock \bibinfo{journal}{\emph{arXiv preprint arXiv:2209.07919}} (\bibinfo{year}{2022}).
\newblock


\bibitem[Moenne-Loccoz et~al\mbox{.}(2024)]%
        {loccoz20243dgrt}
\bibfield{author}{\bibinfo{person}{Nicolas Moenne-Loccoz}, \bibinfo{person}{Ashkan Mirzaei}, \bibinfo{person}{Or Perel}, \bibinfo{person}{Riccardo de Lutio}, \bibinfo{person}{Janick~Martinez Esturo}, \bibinfo{person}{Gavriel State}, \bibinfo{person}{Sanja Fidler}, \bibinfo{person}{Nicholas Sharp}, {and} \bibinfo{person}{Zan Gojcic}.} \bibinfo{year}{2024}\natexlab{}.
\newblock \showarticletitle{3D Gaussian Ray Tracing: Fast Tracing of Particle Scenes}.
\newblock \bibinfo{journal}{\emph{ACM Transactions on Graphics and SIGGRAPH Asia}} (\bibinfo{year}{2024}).
\newblock


\bibitem[Moulon et~al\mbox{.}(2012)]%
        {Moulon2012}
\bibfield{author}{\bibinfo{person}{Pierre Moulon}, \bibinfo{person}{Pascal Monasse}, {and} \bibinfo{person}{Renaud Marlet}.} \bibinfo{year}{2012}\natexlab{}.
\newblock \showarticletitle{Adaptive Structure from Motion with a~Contrario Model Estimation}. In \bibinfo{booktitle}{\emph{Proceedings of the Asian Computer Vision Conference (ACCV 2012)}}. \bibinfo{publisher}{Springer Berlin Heidelberg}, \bibinfo{pages}{257--270}.
\newblock
\href{https://doi.org/10.1007/978-3-642-37447-0_20}{doi:\nolinkurl{10.1007/978-3-642-37447-0_20}}


\bibitem[M{\"u}ller et~al\mbox{.}(2022)]%
        {muller2022instant}
\bibfield{author}{\bibinfo{person}{Thomas M{\"u}ller}, \bibinfo{person}{Alex Evans}, \bibinfo{person}{Christoph Schied}, {and} \bibinfo{person}{Alexander Keller}.} \bibinfo{year}{2022}\natexlab{}.
\newblock \showarticletitle{Instant neural graphics primitives with a multiresolution hash encoding}.
\newblock \bibinfo{journal}{\emph{ACM transactions on graphics (TOG)}} \bibinfo{volume}{41}, \bibinfo{number}{4} (\bibinfo{year}{2022}), \bibinfo{pages}{1--15}.
\newblock


\bibitem[Newcombe et~al\mbox{.}(2011)]%
        {newcombe2011kinectfusion}
\bibfield{author}{\bibinfo{person}{Richard~A Newcombe}, \bibinfo{person}{Shahram Izadi}, \bibinfo{person}{Otmar Hilliges}, \bibinfo{person}{David Molyneaux}, \bibinfo{person}{David Kim}, \bibinfo{person}{Andrew~J Davison}, \bibinfo{person}{Pushmeet Kohi}, \bibinfo{person}{Jamie Shotton}, \bibinfo{person}{Steve Hodges}, {and} \bibinfo{person}{Andrew Fitzgibbon}.} \bibinfo{year}{2011}\natexlab{}.
\newblock \showarticletitle{Kinectfusion: Real-time dense surface mapping and tracking}. In \bibinfo{booktitle}{\emph{2011 10th IEEE international symposium on mixed and augmented reality}}. Ieee, \bibinfo{pages}{127--136}.
\newblock


\bibitem[Park et~al\mbox{.}(2019)]%
        {Park:2019:Deepsdf}
\bibfield{author}{\bibinfo{person}{Jeong~Joon Park}, \bibinfo{person}{Peter Florence}, \bibinfo{person}{Julian Straub}, \bibinfo{person}{Richard~A. Newcombe}, {and} \bibinfo{person}{Steven Lovegrove}.} \bibinfo{year}{2019}\natexlab{}.
\newblock \showarticletitle{{DeepSDF}: Learning Continuous Signed Distance Functions for Shape Representation}. In \bibinfo{booktitle}{\emph{{IEEE} Conference on Computer Vision and Pattern Recognition}}. \bibinfo{pages}{165--174}.
\newblock


\bibitem[Poole et~al\mbox{.}(2022)]%
        {poole2022dreamfusion}
\bibfield{author}{\bibinfo{person}{Ben Poole}, \bibinfo{person}{Ajay Jain}, \bibinfo{person}{Jonathan~T Barron}, {and} \bibinfo{person}{Ben Mildenhall}.} \bibinfo{year}{2022}\natexlab{}.
\newblock \showarticletitle{Dreamfusion: Text-to-3d using 2d diffusion}.
\newblock \bibinfo{journal}{\emph{arXiv preprint arXiv:2209.14988}} (\bibinfo{year}{2022}).
\newblock


\bibitem[Ravi et~al\mbox{.}(2024)]%
        {ravi2024sam2segmentimages}
\bibfield{author}{\bibinfo{person}{Nikhila Ravi}, \bibinfo{person}{Valentin Gabeur}, \bibinfo{person}{Yuan-Ting Hu}, \bibinfo{person}{Ronghang Hu}, \bibinfo{person}{Chaitanya Ryali}, \bibinfo{person}{Tengyu Ma}, \bibinfo{person}{Haitham Khedr}, \bibinfo{person}{Roman Rädle}, \bibinfo{person}{Chloe Rolland}, \bibinfo{person}{Laura Gustafson}, \bibinfo{person}{Eric Mintun}, \bibinfo{person}{Junting Pan}, \bibinfo{person}{Kalyan~Vasudev Alwala}, \bibinfo{person}{Nicolas Carion}, \bibinfo{person}{Chao-Yuan Wu}, \bibinfo{person}{Ross Girshick}, \bibinfo{person}{Piotr Dollár}, {and} \bibinfo{person}{Christoph Feichtenhofer}.} \bibinfo{year}{2024}\natexlab{}.
\newblock \bibinfo{title}{SAM 2: Segment Anything in Images and Videos}.
\newblock
\showeprint[arxiv]{2408.00714}~[cs.CV]
\urldef\tempurl%
\url{https://arxiv.org/abs/2408.00714}
\showURL{%
\tempurl}


\bibitem[Ren et~al\mbox{.}(2024a)]%
        {ren2024grounding}
\bibfield{author}{\bibinfo{person}{Tianhe Ren}, \bibinfo{person}{Qing Jiang}, \bibinfo{person}{Shilong Liu}, \bibinfo{person}{Zhaoyang Zeng}, \bibinfo{person}{Wenlong Liu}, \bibinfo{person}{Han Gao}, \bibinfo{person}{Hongjie Huang}, \bibinfo{person}{Zhengyu Ma}, \bibinfo{person}{Xiaoke Jiang}, \bibinfo{person}{Yihao Chen}, \bibinfo{person}{Yuda Xiong}, \bibinfo{person}{Hao Zhang}, \bibinfo{person}{Feng Li}, \bibinfo{person}{Peijun Tang}, \bibinfo{person}{Kent Yu}, {and} \bibinfo{person}{Lei Zhang}.} \bibinfo{year}{2024}\natexlab{a}.
\newblock \bibinfo{title}{Grounding DINO 1.5: Advance the "Edge" of Open-Set Object Detection}.
\newblock
\showeprint[arxiv]{2405.10300}~[cs.CV]


\bibitem[Ren et~al\mbox{.}(2024b)]%
        {ren2024grounded}
\bibfield{author}{\bibinfo{person}{Tianhe Ren}, \bibinfo{person}{Shilong Liu}, \bibinfo{person}{Ailing Zeng}, \bibinfo{person}{Jing Lin}, \bibinfo{person}{Kunchang Li}, \bibinfo{person}{He Cao}, \bibinfo{person}{Jiayu Chen}, \bibinfo{person}{Xinyu Huang}, \bibinfo{person}{Yukang Chen}, \bibinfo{person}{Feng Yan}, \bibinfo{person}{Zhaoyang Zeng}, \bibinfo{person}{Hao Zhang}, \bibinfo{person}{Feng Li}, \bibinfo{person}{Jie Yang}, \bibinfo{person}{Hongyang Li}, \bibinfo{person}{Qing Jiang}, {and} \bibinfo{person}{Lei Zhang}.} \bibinfo{year}{2024}\natexlab{b}.
\newblock \bibinfo{title}{Grounded SAM: Assembling Open-World Models for Diverse Visual Tasks}.
\newblock
\showeprint[arxiv]{2401.14159}~[cs.CV]


\bibitem[Sarlin et~al\mbox{.}(2019)]%
        {sarlin2019coarse}
\bibfield{author}{\bibinfo{person}{Paul-Edouard Sarlin}, \bibinfo{person}{Cesar Cadena}, \bibinfo{person}{Roland Siegwart}, {and} \bibinfo{person}{Marcin Dymczyk}.} \bibinfo{year}{2019}\natexlab{}.
\newblock \showarticletitle{From Coarse to Fine: Robust Hierarchical Localization at Large Scale}. In \bibinfo{booktitle}{\emph{CVPR}}.
\newblock


\bibitem[Sch{\"o}nberger and Frahm(2016)]%
        {schonberger2016structure}
\bibfield{author}{\bibinfo{person}{Johannes~Lutz Sch{\"o}nberger} {and} \bibinfo{person}{Jan-Michael Frahm}.} \bibinfo{year}{2016}\natexlab{}.
\newblock \showarticletitle{Structure-from-motion revisited}. In \bibinfo{booktitle}{\emph{IEEE Conference on Computer Vision and Pattern Recognition (CVPR)}}. \bibinfo{pages}{4104--4113}.
\newblock


\bibitem[Sch\"{o}nberger et~al\mbox{.}(2016)]%
        {schoenberger2016mvs}
\bibfield{author}{\bibinfo{person}{Johannes~Lutz Sch\"{o}nberger}, \bibinfo{person}{Enliang Zheng}, \bibinfo{person}{Marc Pollefeys}, {and} \bibinfo{person}{Jan-Michael Frahm}.} \bibinfo{year}{2016}\natexlab{}.
\newblock \showarticletitle{Pixelwise View Selection for Unstructured Multi-View Stereo}. In \bibinfo{booktitle}{\emph{European Conference on Computer Vision (ECCV)}}.
\newblock


\bibitem[Sun et~al\mbox{.}(2024)]%
        {sun2024nu-nerf}
\bibfield{author}{\bibinfo{person}{Jia-Mu Sun}, \bibinfo{person}{Tong Wu}, \bibinfo{person}{Ling-Qi Yan}, {and} \bibinfo{person}{Lin Gao}.} \bibinfo{year}{2024}\natexlab{}.
\newblock \showarticletitle{NU-NeRF: Neural Reconstruction of Nested Transparent Objects with Uncontrolled Capture Environment}.
\newblock \bibinfo{journal}{\emph{ACM Trans. Graph.}} \bibinfo{volume}{43}, \bibinfo{number}{6}, Article \bibinfo{articleno}{262} (\bibinfo{date}{Nov.} \bibinfo{year}{2024}), \bibinfo{numpages}{14}~pages.
\newblock
\showISSN{0730-0301}
\href{https://doi.org/10.1145/3687757}{doi:\nolinkurl{10.1145/3687757}}


\bibitem[Tang et~al\mbox{.}(2023)]%
        {tang2023dreamgaussian}
\bibfield{author}{\bibinfo{person}{Jiaxiang Tang}, \bibinfo{person}{Jiawei Ren}, \bibinfo{person}{Hang Zhou}, \bibinfo{person}{Ziwei Liu}, {and} \bibinfo{person}{Gang Zeng}.} \bibinfo{year}{2023}\natexlab{}.
\newblock \showarticletitle{Dreamgaussian: Generative gaussian splatting for efficient 3d content creation}.
\newblock \bibinfo{journal}{\emph{arXiv preprint arXiv:2309.16653}} (\bibinfo{year}{2023}).
\newblock


\bibitem[Wang et~al\mbox{.}(2023)]%
        {Wang_2023_ICCV}
\bibfield{author}{\bibinfo{person}{Dongqing Wang}, \bibinfo{person}{Tong Zhang}, {and} \bibinfo{person}{Sabine S\"usstrunk}.} \bibinfo{year}{2023}\natexlab{}.
\newblock \showarticletitle{NEMTO: Neural Environment Matting for Novel View and Relighting Synthesis of Transparent Objects}. In \bibinfo{booktitle}{\emph{Proceedings of the IEEE/CVF International Conference on Computer Vision (ICCV)}}. \bibinfo{pages}{317--327}.
\newblock


\bibitem[Wang et~al\mbox{.}(2021a)]%
        {wang2021patchmatchnet}
\bibfield{author}{\bibinfo{person}{Fangjinhua Wang}, \bibinfo{person}{Silvano Galliani}, \bibinfo{person}{Christoph Vogel}, \bibinfo{person}{Pablo Speciale}, {and} \bibinfo{person}{Marc Pollefeys}.} \bibinfo{year}{2021}\natexlab{a}.
\newblock \showarticletitle{Patchmatchnet: Learned multi-view patchmatch stereo}. In \bibinfo{booktitle}{\emph{Proceedings of the IEEE/CVF conference on computer vision and pattern recognition}}. \bibinfo{pages}{14194--14203}.
\newblock


\bibitem[Wang et~al\mbox{.}(2018)]%
        {Wang:2018:Pixel2mesh}
\bibfield{author}{\bibinfo{person}{Nanyang Wang}, \bibinfo{person}{Yinda Zhang}, \bibinfo{person}{Zhuwen Li}, \bibinfo{person}{Yanwei Fu}, \bibinfo{person}{Wei Liu}, {and} \bibinfo{person}{Yu{-}Gang Jiang}.} \bibinfo{year}{2018}\natexlab{}.
\newblock \showarticletitle{Pixel2Mesh: {Generating} {3D} Mesh Models from Single {RGB} Images}. In \bibinfo{booktitle}{\emph{European Conference on Computer Vision}}, Vol.~\bibinfo{volume}{11215}. \bibinfo{pages}{55--71}.
\newblock


\bibitem[Wang et~al\mbox{.}(2021b)]%
        {wang2021neus}
\bibfield{author}{\bibinfo{person}{Peng Wang}, \bibinfo{person}{Lingjie Liu}, \bibinfo{person}{Yuan Liu}, \bibinfo{person}{Christian Theobalt}, \bibinfo{person}{Taku Komura}, {and} \bibinfo{person}{Wenping Wang}.} \bibinfo{year}{2021}\natexlab{b}.
\newblock \showarticletitle{NeuS: learning neural implicit surfaces by volume rendering for multi-view reconstruction}. In \bibinfo{booktitle}{\emph{Proceedings of the 35th International Conference on Neural Information Processing Systems}} \emph{(\bibinfo{series}{NIPS '21})}. \bibinfo{publisher}{Curran Associates Inc.}, \bibinfo{address}{Red Hook, NY, USA}, Article \bibinfo{articleno}{2081}, \bibinfo{numpages}{13}~pages.
\newblock
\showISBNx{9781713845393}


\bibitem[Wu(2013)]%
        {wu2013towards}
\bibfield{author}{\bibinfo{person}{Changchang Wu}.} \bibinfo{year}{2013}\natexlab{}.
\newblock \showarticletitle{Towards linear-time incremental structure from motion}. In \bibinfo{booktitle}{\emph{2013 International Conference on 3D Vision-3DV 2013}}. IEEE, \bibinfo{pages}{127--134}.
\newblock


\bibitem[Wu et~al\mbox{.}(2025)]%
        {wu20253dgut}
\bibfield{author}{\bibinfo{person}{Qi Wu}, \bibinfo{person}{Janick Martinez~Esturo}, \bibinfo{person}{Ashkan Mirzaei}, \bibinfo{person}{Nicolas Moenne-Loccoz}, {and} \bibinfo{person}{Zan Gojcic}.} \bibinfo{year}{2025}\natexlab{}.
\newblock \showarticletitle{3DGUT: Enabling Distorted Cameras and Secondary Rays in Gaussian Splatting}.
\newblock \bibinfo{journal}{\emph{Conference on Computer Vision and Pattern Recognition (CVPR)}} (\bibinfo{year}{2025}).
\newblock


\bibitem[Wu et~al\mbox{.}({[n.\,d.]})]%
        {wu2025alphasurf}
\bibfield{author}{\bibinfo{person}{Tianhao~Walter Wu}, \bibinfo{person}{Fangcheng Zhong}, \bibinfo{person}{Gernot Riegler}, \bibinfo{person}{Shimon Vainer}, \bibinfo{person}{Jiankang Deng}, \bibinfo{person}{Cengiz Oztireli}, {et~al\mbox{.}}} \bibinfo{year}{[n.\,d.]}\natexlab{}.
\newblock \showarticletitle{$\alpha$surf: Implicit surface reconstruction for semi-transparent and thin objects with decoupled geometry and opacity}. In \bibinfo{booktitle}{\emph{International Conference on 3D Vision 2025}}.
\newblock


\bibitem[Xie et~al\mbox{.}(2019)]%
        {Xie:2019:Pix2vox}
\bibfield{author}{\bibinfo{person}{Haozhe Xie}, \bibinfo{person}{Hongxun Yao}, \bibinfo{person}{Xiaoshuai Sun}, \bibinfo{person}{Shangchen Zhou}, {and} \bibinfo{person}{Shengping Zhang}.} \bibinfo{year}{2019}\natexlab{}.
\newblock \showarticletitle{Pix2{V}ox: Context-Aware {3D} Reconstruction From Single and Multi-View Images}. In \bibinfo{booktitle}{\emph{{IEEE/CVF} International Conference on Computer Vision}}. \bibinfo{pages}{2690--2698}.
\newblock


\bibitem[Xu et~al\mbox{.}(2022)]%
        {xu2022point}
\bibfield{author}{\bibinfo{person}{Qiangeng Xu}, \bibinfo{person}{Zexiang Xu}, \bibinfo{person}{Julien Philip}, \bibinfo{person}{Sai Bi}, \bibinfo{person}{Zhixin Shu}, \bibinfo{person}{Kalyan Sunkavalli}, {and} \bibinfo{person}{Ulrich Neumann}.} \bibinfo{year}{2022}\natexlab{}.
\newblock \showarticletitle{Point-nerf: Point-based neural radiance fields}. In \bibinfo{booktitle}{\emph{Proceedings of the IEEE/CVF conference on computer vision and pattern recognition}}. \bibinfo{pages}{5438--5448}.
\newblock


\bibitem[Yang et~al\mbox{.}(2024)]%
        {yang2024spec}
\bibfield{author}{\bibinfo{person}{Ziyi Yang}, \bibinfo{person}{Xinyu Gao}, \bibinfo{person}{Yangtian Sun}, \bibinfo{person}{Yihua Huang}, \bibinfo{person}{Xiaoyang Lyu}, \bibinfo{person}{Wen Zhou}, \bibinfo{person}{Shaohui Jiao}, \bibinfo{person}{Xiaojuan Qi}, {and} \bibinfo{person}{Xiaogang Jin}.} \bibinfo{year}{2024}\natexlab{}.
\newblock \showarticletitle{Spec-gaussian: Anisotropic view-dependent appearance for 3d gaussian splatting}.
\newblock \bibinfo{journal}{\emph{arXiv preprint arXiv:2402.15870}} (\bibinfo{year}{2024}).
\newblock


\bibitem[Yariv et~al\mbox{.}(2021)]%
        {Yariv:2021:Volume}
\bibfield{author}{\bibinfo{person}{Lior Yariv}, \bibinfo{person}{Jiatao Gu}, \bibinfo{person}{Yoni Kasten}, {and} \bibinfo{person}{Yaron Lipman}.} \bibinfo{year}{2021}\natexlab{}.
\newblock \showarticletitle{Volume Rendering of Neural Implicit Surfaces}. In \bibinfo{booktitle}{\emph{Advances in Neural Information Processing Systems}}. \bibinfo{pages}{4805--4815}.
\newblock


\bibitem[Ye et~al\mbox{.}(2024b)]%
        {ye2024stablenormal}
\bibfield{author}{\bibinfo{person}{Chongjie Ye}, \bibinfo{person}{Lingteng Qiu}, \bibinfo{person}{Xiaodong Gu}, \bibinfo{person}{Qi Zuo}, \bibinfo{person}{Yushuang Wu}, \bibinfo{person}{Zilong Dong}, \bibinfo{person}{Liefeng Bo}, \bibinfo{person}{Yuliang Xiu}, {and} \bibinfo{person}{Xiaoguang Han}.} \bibinfo{year}{2024}\natexlab{b}.
\newblock \showarticletitle{StableNormal: Reducing Diffusion Variance for Stable and Sharp Normal}.
\newblock \bibinfo{journal}{\emph{ACM Transactions on Graphics (TOG)}} (\bibinfo{year}{2024}).
\newblock


\bibitem[Ye et~al\mbox{.}(2023)]%
        {Ye2023IntrinsicNeRF}
\bibfield{author}{\bibinfo{person}{Weicai Ye}, \bibinfo{person}{Shuo Chen}, \bibinfo{person}{Chong Bao}, \bibinfo{person}{Hujun Bao}, \bibinfo{person}{Marc Pollefeys}, \bibinfo{person}{Zhaopeng Cui}, {and} \bibinfo{person}{Guofeng Zhang}.} \bibinfo{year}{2023}\natexlab{}.
\newblock \showarticletitle{{IntrinsicNeRF: Learning Intrinsic Neural Radiance Fields for Editable Novel View Synthesis}}. In \bibinfo{booktitle}{\emph{{Proceedings of the IEEE/CVF International Conference on Computer Vision}}}.
\newblock


\bibitem[Ye et~al\mbox{.}(2021)]%
        {ye2021superplane}
\bibfield{author}{\bibinfo{person}{Weicai Ye}, \bibinfo{person}{Hai Li}, \bibinfo{person}{Tianxiang Zhang}, \bibinfo{person}{Xiaowei Zhou}, \bibinfo{person}{Hujun Bao}, {and} \bibinfo{person}{Guofeng Zhang}.} \bibinfo{year}{2021}\natexlab{}.
\newblock \showarticletitle{{SuperPlane: 3D plane detection and description from a single image}}. In \bibinfo{booktitle}{\emph{2021 IEEE Virtual Reality and 3D User Interfaces (VR)}}. IEEE, \bibinfo{pages}{207--215}.
\newblock


\bibitem[Ye et~al\mbox{.}(2024a)]%
        {ye2024absgs}
\bibfield{author}{\bibinfo{person}{Zongxin Ye}, \bibinfo{person}{Wenyu Li}, \bibinfo{person}{Sidun Liu}, \bibinfo{person}{Peng Qiao}, {and} \bibinfo{person}{Yong Dou}.} \bibinfo{year}{2024}\natexlab{a}.
\newblock \showarticletitle{AbsGS: Recovering Fine Details in 3D Gaussian Splatting}. In \bibinfo{booktitle}{\emph{Proceedings of the 32nd ACM International Conference on Multimedia}} (Melbourne VIC, Australia) \emph{(\bibinfo{series}{MM '24})}. \bibinfo{publisher}{Association for Computing Machinery}, \bibinfo{address}{New York, NY, USA}, \bibinfo{pages}{1053–1061}.
\newblock
\showISBNx{9798400706868}
\href{https://doi.org/10.1145/3664647.3681361}{doi:\nolinkurl{10.1145/3664647.3681361}}


\bibitem[Yu et~al\mbox{.}(2024)]%
        {yu2024gaussian}
\bibfield{author}{\bibinfo{person}{Zehao Yu}, \bibinfo{person}{Torsten Sattler}, {and} \bibinfo{person}{Andreas Geiger}.} \bibinfo{year}{2024}\natexlab{}.
\newblock \showarticletitle{Gaussian Opacity Fields: Efficient and Compact Surface Reconstruction in Unbounded Scenes}.
\newblock \bibinfo{journal}{\emph{arXiv preprint arXiv:2404.10772}} (\bibinfo{year}{2024}).
\newblock


\bibitem[Zhang et~al\mbox{.}(2025)]%
        {zhang2024from}
\bibfield{author}{\bibinfo{person}{Haoran Zhang}, \bibinfo{person}{Junkai Deng}, \bibinfo{person}{Xuhui Chen}, \bibinfo{person}{Fei Hou}, \bibinfo{person}{Wencheng Wang}, \bibinfo{person}{Hong Qin}, \bibinfo{person}{Chen Qian}, {and} \bibinfo{person}{Ying He}.} \bibinfo{year}{2025}\natexlab{}.
\newblock \showarticletitle{From transparent to opaque: rethinking neural implicit surfaces with $\alpha$-NeuS}. In \bibinfo{booktitle}{\emph{Proceedings of the 38th International Conference on Neural Information Processing Systems}} (Vancouver, BC, Canada) \emph{(\bibinfo{series}{NIPS '24})}. \bibinfo{publisher}{Curran Associates Inc.}, \bibinfo{address}{Red Hook, NY, USA}, Article \bibinfo{articleno}{880}, \bibinfo{numpages}{25}~pages.
\newblock
\showISBNx{9798331314385}


\bibitem[Zhang et~al\mbox{.}(2018)]%
        {zhang2018unreasonable}
\bibfield{author}{\bibinfo{person}{Richard Zhang}, \bibinfo{person}{Phillip Isola}, \bibinfo{person}{Alexei~A Efros}, \bibinfo{person}{Eli Shechtman}, {and} \bibinfo{person}{Oliver Wang}.} \bibinfo{year}{2018}\natexlab{}.
\newblock \showarticletitle{The unreasonable effectiveness of deep features as a perceptual metric}. In \bibinfo{booktitle}{\emph{Proceedings of the IEEE conference on computer vision and pattern recognition}}. \bibinfo{pages}{586--595}.
\newblock


\bibitem[Zhang et~al\mbox{.}(1999)]%
        {zhang1999shape}
\bibfield{author}{\bibinfo{person}{Ruo Zhang}, \bibinfo{person}{Ping-Sing Tsai}, \bibinfo{person}{J.E. Cryer}, {and} \bibinfo{person}{M. Shah}.} \bibinfo{year}{1999}\natexlab{}.
\newblock \showarticletitle{Shape-from-shading: a survey}.
\newblock \bibinfo{journal}{\emph{IEEE Transactions on Pattern Analysis and Machine Intelligence}} \bibinfo{volume}{21}, \bibinfo{number}{8} (\bibinfo{year}{1999}), \bibinfo{pages}{690--706}.
\newblock
\href{https://doi.org/10.1109/34.784284}{doi:\nolinkurl{10.1109/34.784284}}


\bibitem[Zhao et~al\mbox{.}(2001)]%
        {zhao2001fast}
\bibfield{author}{\bibinfo{person}{Hong-Kai Zhao}, \bibinfo{person}{Stanley Osher}, {and} \bibinfo{person}{Ronald Fedkiw}.} \bibinfo{year}{2001}\natexlab{}.
\newblock \showarticletitle{Fast surface reconstruction using the level set method}. In \bibinfo{booktitle}{\emph{Proceedings of the IEEE Workshop on Variational and Level Set Methods}}. IEEE, \bibinfo{pages}{194--201}.
\newblock


\bibitem[Zhou et~al\mbox{.}(2025)]%
        {deweizhou2025dreamrenderer}
\bibfield{author}{\bibinfo{person}{Dewei Zhou}, \bibinfo{person}{Mingwei Li}, \bibinfo{person}{Zongxin Yang}, {and} \bibinfo{person}{Yi Yang}.} \bibinfo{year}{2025}\natexlab{}.
\newblock \bibinfo{title}{DreamRenderer: Taming Multi-Instance Attribute Control in Large-Scale Text-to-Image Models}.
\newblock
\showeprint[arxiv]{2503.12885}~[cs.CV]
\urldef\tempurl%
\url{https://arxiv.org/abs/2503.12885}
\showURL{%
\tempurl}


\end{thebibliography}
% 不含appendix
\appendix
\clearpage
\section{Appendix}
\label{sec:appendix}

\subsection{TransLab Dataset}
\label{subsec:tlab-dataset}

The rise of embodied AI in automated laboratories necessitates accurate 3D reconstruction and manipulation of transparent glassware. To address the lack of specialized benchmarks for this challenging task, the TransLab dataset provides a comprehensive synthetic benchmark for evaluating reconstruction algorithms on transparent objects. It features a diverse collection of 8 scenes in total, containing typical laboratory glassware, including test tubes, beakers, safety goggles, flasks (two-neck and three-neck variants), petri dishes, graduated cylinders, round-bottom flasks, Erlenmeyer flasks, and condensers.

For each scene, we provide extensive ground truth data, generated using Blender's physically-based rendering (PBR) engine, to facilitate thorough evaluation. This includes high-resolution RGB imagery (\(I_{GT}\)) rendered at an original resolution of 1600x1600 pixels (downscaled to 800x800 for training), reference 3D ground truth meshes (\(\mathcal{M}_{GT}\)), background segmentation masks (\(M_{BG}\)), masks identifying transparent objects (\(M_{T}\)), dense depth maps (\(D\)), surface normal maps (\(N\)), and environmental illumination maps (\(E\)). A summary of these data types is also provided in Table~\ref{tab:tlab_data}. This rich set of annotations enables detailed analysis of reconstruction quality from multiple perspectives.

\begin{table}[h]
	\centering
	\caption{Ground Truth Data Provided in the TransLab Dataset}
	\label{tab:tlab_data}
	\begin{tabular}{cl}
		\toprule
		Symbol               & Description                           \\
		\midrule
		\(I_{GT}\)           & High-resolution RGB imagery           \\
		\(\mathcal{M}_{GT}\) & Reference 3D ground truth meshes      \\
		\(M_{BG}\)           & Background segmentation masks         \\
		\(M_{T}\)            & Transparent object segmentation masks \\
		\(D\)                & Dense depth maps                      \\
		\(N\)                & Surface normal maps                   \\
		\(E\)                & Environmental illumination maps       \\
		\bottomrule
	\end{tabular}
\end{table}

% \subsection{ClearPose dataset}

\subsection{Quantitative Evaluation Details}
To quantitatively assess reconstruction performance, we employ established metrics for both image quality and geometric fidelity. Image quality is measured using PSNR, SSIM, and LPIPS. Geometric accuracy is evaluated using two primary metrics:

% 下面这段，计算CD过程，是从Mesh里获取一个point set然后计算吗？
\paragraph{Chamfer Distance (CD)} This metric quantifies the average distance between the predicted mesh surface \(\mathcal{M}_{\text{pred}}\) and the ground truth mesh surface \(\mathcal{M}_{\text{GT}}\). To compute the CD, we first uniformly sample a dense set of points \(P_{\text{pred}}\) from the surface of the predicted mesh \(\mathcal{M}_{\text{pred}}\) and similarly sample a point set \(P_{\text{GT}}\) from the ground truth mesh \(\mathcal{M}_{\text{GT}}\). The chamfer distance is then calculated as the symmetric average of the mean shortest distances between the points in these two point clouds.

\paragraph{Geometry F1 Score} This metric assesses the balance between the precision and recall of the reconstructed mesh vertices. Let \(V_{\text{pred}}\) be the set of vertices in the predicted mesh \(\mathcal{M}_{\text{pred}}\) and \(V_{\text{GT}}\) be the set of vertices in the ground truth mesh \(\mathcal{M}_{\text{GT}}\). A predicted vertex \(v \in V_{\text{pred}}\) is considered a true positive (TP) if its Euclidean distance to the nearest ground truth vertex \(v' \in V_{\text{GT}}\) is less than a predefined threshold \(\tau = 0.005\):
\begin{equation}
	TP = \{v \in V_{\text{pred}} \mid \min_{v' \in V_{\text{GT}}} \|v - v'\|_2 < \tau\}
\end{equation}
Precision (\(P\)) is defined as the ratio of true positives to the total number of predicted vertices, \(P = | TP | / | V_{\text{pred}} |\). Recall (\(R\)) is defined as the ratio of true positives to the total number of ground truth vertices, \(R = | TP | / | V_{\text{GT}} |\). The F1 score is the harmonic mean of precision and recall:
\begin{equation}
	F1 = 2 \cdot \frac{P \cdot R}{P + R}
\end{equation}

\subsection{Error Analysis of Depth Extraction}

We set the depth window size \(\delta t\) to 3\,mm in our maximum-weight window search. Since the final estimated depth \(d_{\text{first}}\) is computed as a weighted average within the selected window \(\mathcal{W}^*\), the theoretical depth error is naturally bounded by the window size, i.e., at most \(\delta t = 3\)mm.

\subsection{Limitations and Future Work}
\noindent\textbf{Limitations.} Our framework currently assumes single-layer transparency; the first-surface depth extraction may be less accurate for complex multi-layer refractive phenomena (e.g., liquids in containers) leading to caustics. Additionally, modeling highly complex or spatially varying anisotropic material appearances remains an avenue for improvement.

\noindent\textbf{Future work.} Future work can directly address these limitations. Key directions include: (1)~Extending our first-surface extraction method to handle multi-layer transparencies and complex refractive effects more accurately, potentially by integrating principles from physically-based ray tracing. (2)~Improving the modeling of complex, spatially varying material appearances by leveraging advanced neural BRDF representations or integrating material spectroscopy priors. These advancements promise to further enhance the accuracy and realism of transparent object reconstruction. % Rewritten future work based on limitations

\subsection{Additional Qualitative Results}
\label{subsec:additional-qualitative-results}

In this section, we present additional qualitative results to further demonstrate the performance of our method, TSGS, on both the challenging TransLab dataset and the general DTU benchmark.

\subsubsection{TransLab Dataset Results}
\begin{figure*}[ht!]
    \centering
    \includegraphics[width=\linewidth]{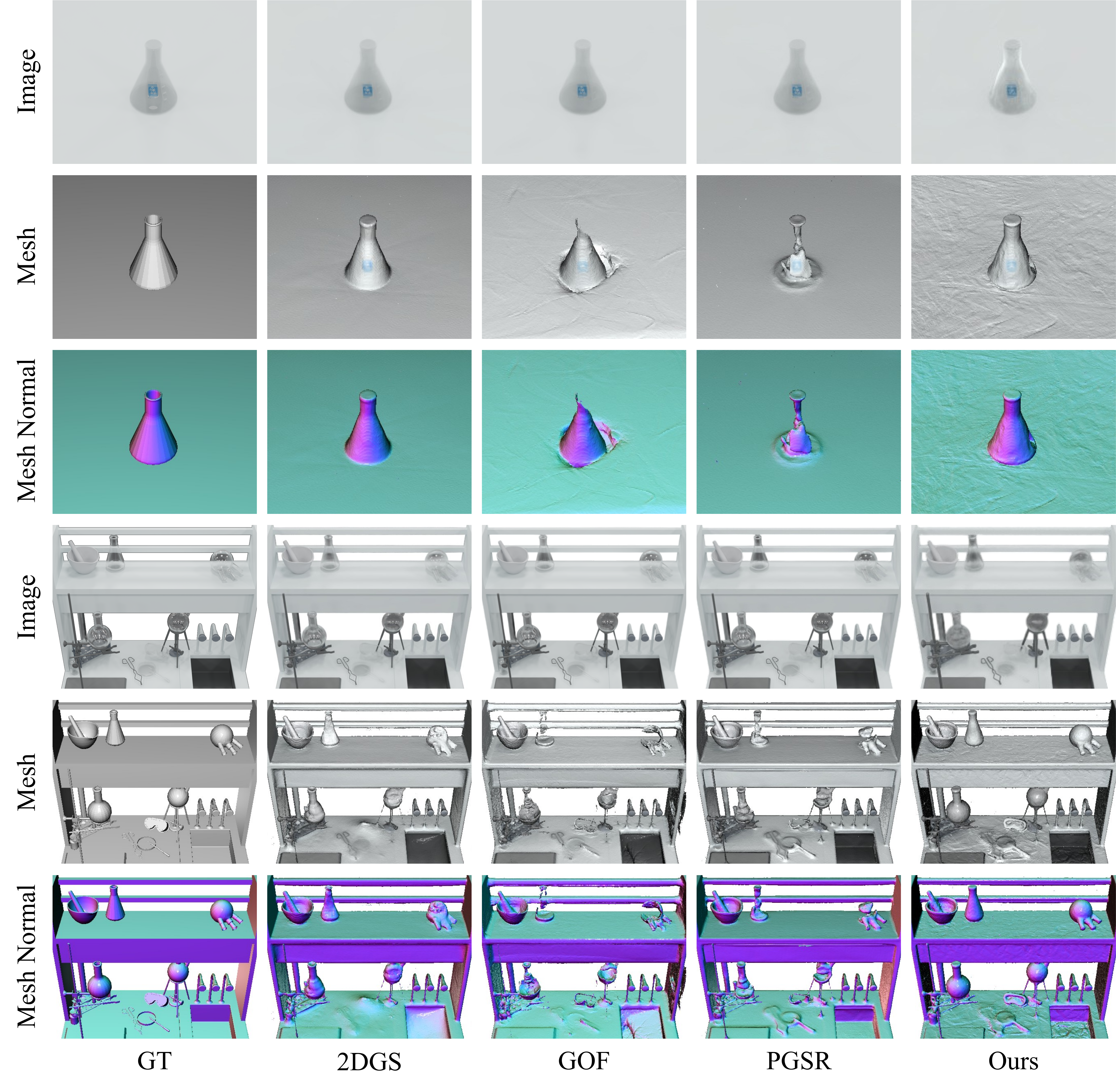}
    \caption{Additional qualitative reconstruction results on various scenes from the TransLab dataset. Our method (TSGS) successfully captures the geometry and appearance of complex transparent objects like test tubes, beakers, and flasks.}
    \label{fig:additional_qualitative_translab_1}
\end{figure*}

\begin{figure*}[p!] 
    \centering
    % \vspace*{\fill} % 添加第一个垂直填充
    \includegraphics[width=\linewidth]{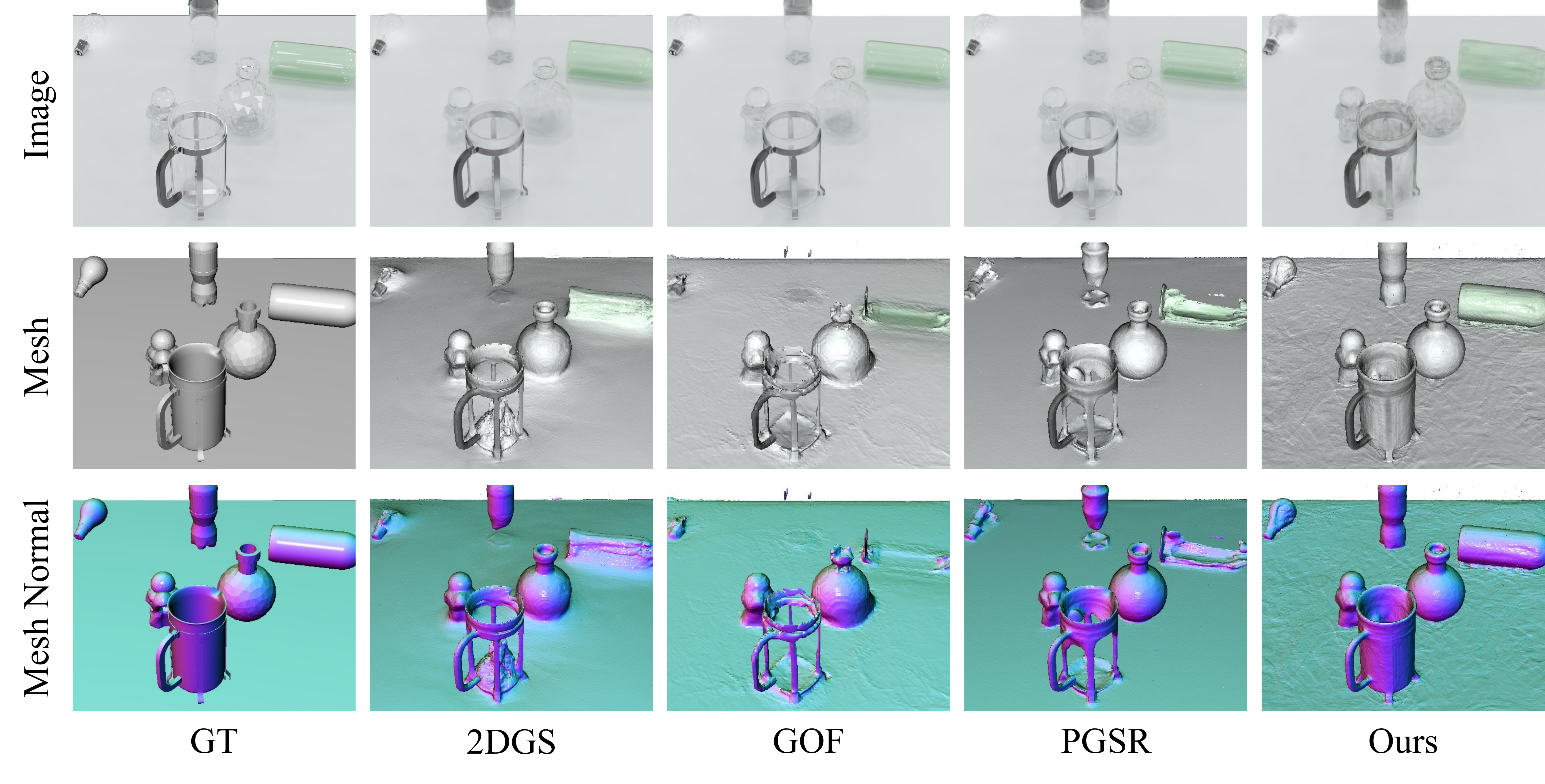} % Placeholder image path
    % \vspace*{\fill} % 添加第二个垂直填充
    \caption{Additional qualitative reconstruction results on various scenes from the TransLab dataset. Our method (TSGS) successfully captures the geometry and appearance of complex transparent objects like test tubes, beakers, and flasks.}
    \label{fig:additional_qualitative_translab_2} % Make sure label is unique
\end{figure*}

Fig.~\ref{fig:additional_qualitative_translab_1} and Fig.~\ref{fig:additional_qualitative_translab_2} showcases more reconstruction examples from the TransLab dataset. These results highlight TSGS's ability to handle diverse transparent laboratory equipment with varying geometric complexity and material properties. Our method consistently recovers accurate first-surface geometry while rendering photorealistic appearances, effectively addressing the challenges posed by transparency.
\subsubsection{DTU Dataset Results}
To illustrate the generalizability of our approach, Fig.~\ref{fig:additional_qualitative_dtu_1}, Fig.~\ref{fig:additional_qualitative_dtu_2}, Fig.~\ref{fig:additional_qualitative_dtu_3}, Fig.~\ref{fig:additional_qualitative_dtu_4}, Fig.~\ref{fig:additional_qualitative_dtu_5} present additional qualitative results on selected scenes from the DTU dataset. Although optimized for transparent surfaces, TSGS demonstrates strong performance on these standard opaque objects, achieving high-fidelity surface reconstruction comparable to state-of-the-art methods designed for general scenes. This underscores the robustness of our geometry learning stage. For the results of PGSR, 2DGS, and GOF, we directly adopt the visualizations from the original PGSR paper, please note that minor viewpoint discrepancies may exist as we did not have access to their original camera and projection parameters.

\subsection{Additional Quantitative Results}

\subsubsection{Results on Real-world Dataset}
To validate our method's performance in real-world scenarios, we conduct a quantitative evaluation on the highly challenging ClearPose~\cite{chen2022clearpose} dataset. This benchmark features 63 scenes with highly transparent objects, which often exhibit complex overlaps and out-of-focus blur. For this evaluation, we selected ``scene1'' from each of the four sets. Given the high density of views in the original dataset, we subsampled the training data by selecting one image every 100 frames. The quantitative results are presented in Table~\ref{tab:clearpose}. Note that the ClearPose dataset provides ground truth (GT) only for the target objects by aligning a mesh with the object in the captured images. Consequently, GT is not available for other parts of the scene, such as the table or background. Thus, for geometric evaluation, we compute a unidirectional Chamfer Distance from the predicted mesh to the GT object mesh.
\begin{table*}[t]
	\caption{Quantitative evaluation on the ClearPose~\cite{chen2022clearpose} dataset. For each scene, we compare our method against 2DGS~\cite{Huang2DGS2024}, PGSR~\cite{chen2024pgsr}, and NU-NeRF~\cite{sun2024nu-nerf}. Best results are in \textbf{bold}.}
	\label{tab:clearpose}
	\centering
	\setlength{\tabcolsep}{6pt} % Adjust column spacing
	\begin{tabular}{@{}llccccc@{}}
		\toprule
		Scene & Method                        & PSNR $\uparrow$ & SSIM $\uparrow$ & LPIPS $\downarrow$ & CD $\downarrow$ & F1 $\uparrow$  \\
		\midrule
		\multicolumn{7}{@{}l}{\textit{\textbf{Chemical}}}                                                                                 \\
		\midrule
		Set 1 & 2DGS~\cite{Huang2DGS2024}     & 21.20           & 70.6            & 0.447              & 30.03           & 5.35           \\
		      & PGSR~\cite{chen2024pgsr}      & 21.18           & 78.0            & 0.384              & 15.28           & 22.37          \\
		      & NU-NeRF~\cite{sun2024nu-nerf} & 20.53           & 73.3            & 0.440              & 27.88           & 7.45           \\
		      & Ours                          & \textbf{22.20}  & \textbf{77.3}   & \textbf{0.348}     & \textbf{7.73}   & \textbf{27.56} \\
		\midrule
		\multicolumn{7}{@{}l}{\textit{\textbf{Household}}}                                                                                \\
		\midrule
		Set 2 & 2DGS~\cite{Huang2DGS2024}     & 16.85           & 40.8            & 0.499              & 36.15           & 8.22           \\
		      & PGSR~\cite{chen2024pgsr}      & 13.39           & 27.0            & 0.574              & 20.92           & 7.00           \\
		      & NU-NeRF~\cite{sun2024nu-nerf} & 16.99           & 49.7            & 0.503              & 12.02           & 36.44          \\
		      & Ours                          & \textbf{17.90}  & \textbf{54.0}   & \textbf{0.410}     & \textbf{7.48}   & \textbf{36.53} \\
		\midrule
		Set 3 & 2DGS~\cite{Huang2DGS2024}     & 15.30           & 40.7            & 0.524              & 45.77           & 17.10          \\
		      & PGSR~\cite{chen2024pgsr}      & 13.65           & 25.7            & 0.561              & 13.88           & 4.06           \\
		      & NU-NeRF~\cite{sun2024nu-nerf} & 16.23           & 53.6            & 0.501              & 10.55           & 37.47          \\
		      & Ours                          & \textbf{16.89}  & \textbf{54.4}   & \textbf{0.412}     & \textbf{6.54}   & \textbf{42.09} \\
		\midrule
		Set 4 & 2DGS~\cite{Huang2DGS2024}     & 14.50           & 37.8            & 0.543              & 54.48           & 12.67          \\
		      & PGSR~\cite{chen2024pgsr}      & 14.52           & 29.3            & 0.580              & 11.36           & 24.06          \\
		      & NU-NeRF~\cite{sun2024nu-nerf} & \textbf{17.37}  & \textbf{62.8}   & \textbf{0.386}     & 17.35           & 35.13          \\
		      & Ours                          & 15.90           & 57.4            & 0.393              & \textbf{7.89}   & \textbf{36.86} \\
		\bottomrule
	\end{tabular}
\end{table*}

\subsubsection{Computational Efficiency}
\label{sec:efficiency}

As shown in Table~\ref{tab:efficiency}, our 3DGS-based method is significantly faster than NeRF-based approaches and exhibits performance comparable to other 3DGS-based methods. Specifically, our method's training time of 0.8 hours is substantially lower than the 13 hours required by NU-NeRF. It is worth noting that due to its requirement for NVIDIA OptiX, the NU-NeRF results were obtained on a machine equipped with an NVIDIA RTX 4090 GPU. In terms of rendering speed, our approach achieves 105 FPS, which is on par with 2DGS and faster than PGSR. This efficient performance, combined with moderate VRAM usage, underscores our method's suitability for applications requiring rapid 3D reconstruction.

\begin{table*}[t]
	\centering
	\caption{Computational efficiency comparison. Our method is significantly faster than NeRF-based approaches and comparable to other 3DGS-based methods.}
	\label{tab:efficiency}
	\begin{tabular}{@{}lccc@{}}
		\toprule
		Method & Training Time (hour) & VRAM (GB) & Render FPS \\
		\midrule
		2DGS~\cite{Huang2DGS2024}      & 0.2                & 2         & 108        \\
		PGSR~\cite{chen2024pgsr}      & 0.5                & 6         & 87         \\
		NU-NeRF~\cite{sun2024nu-nerf}   & 13                 & 20        & 0.03       \\
		Ours      & 0.8                & 7.7       & 105        \\
		\bottomrule
	\end{tabular}
\end{table*}

\subsubsection{Comparison with Baselines Using Normal Priors}
\label{sec:comparison_with_normal_priors}

To further evaluate our method, we provide a comparison against baseline methods enhanced with the same normal priors on the TransLab dataset. As shown in Table~\ref{tab:translab_with_normal}, our approach consistently outperforms the enhanced baselines across key metrics. This demonstrates that the superior performance of our method is attributed to its fundamental design, rather than being solely dependent on the auxiliary normal information.

\begin{table*}[t]
	\centering
	\caption{Comparison on the TransLab dataset with baseline methods enhanced with normal priors. Best results are in \textbf{bold}.}
	\label{tab:translab_with_normal}
	\begin{tabular}{@{}lccccc@{}}
		\toprule
		Method        & PSNR $\uparrow$ & SSIM $\uparrow$ & LPIPS $\downarrow$ & CD $\downarrow$ & F1 $\uparrow$ \\
		\midrule
		2DGS+Normal   & 29.68           & 97.7            & 0.040              & 2.09            & 92.6          \\
		PGSR+Normal   & 38.22           & 98.6            & \textbf{0.020}     & 2.20            & 91.2          \\
		Ours          & \textbf{39.08}  & \textbf{98.9}   & \textbf{0.020}     & \textbf{1.85}   & \textbf{95.0} \\
		\bottomrule
	\end{tabular}
\end{table*}

\subsubsection{Comparison with NU-NeRF on the TransLab Dataset}
\label{sec:comparison_nunerf_translab}

We also conducted a direct comparison against NU-NeRF on the TransLab dataset. A significant challenge observed with NU-NeRF is its tendency to suffer from convergence issues on this dataset, often resulting in degenerated reconstructions that collapse into a simple spherical shape. In contrast, our method demonstrates robust convergence and superior reconstruction quality. The quantitative comparison of the average metrics is presented in Table~\ref{tab:comparison_nunerf}, which clearly highlights the performance gap between the two methods.

\begin{table*}[t]
	\centering
	\caption{Quantitative comparison with NU-NeRF on the TransLab dataset. We report the average metrics across all scenes. Best results are in \textbf{bold}.}
	\label{tab:comparison_nunerf}
	\begin{tabular}{@{}lccccc@{}}
		\toprule
		Method  & PSNR $\uparrow$ & SSIM $\uparrow$ & LPIPS $\downarrow$ & CD $\downarrow$ & F1 $\uparrow$ \\
		\midrule
		NU-NeRF & 28.20           & 92.9            & 0.144              & 34.68           & 36.0          \\
		Ours    & \textbf{39.08}  & \textbf{98.9}   & \textbf{0.020}     & \textbf{1.85}   & \textbf{95.0} \\
		\bottomrule
	\end{tabular}
\end{table*}


\newlength{\mydtufigureheight}
\setlength{\mydtufigureheight}{18cm} % 设置你想要的统一高度，例如8cm

\begin{figure*}[htbp!]
	\centering
	\vspace*{\fill} % 添加第一个垂直填充
	\includegraphics[height=\mydtufigureheight, keepaspectratio=true]{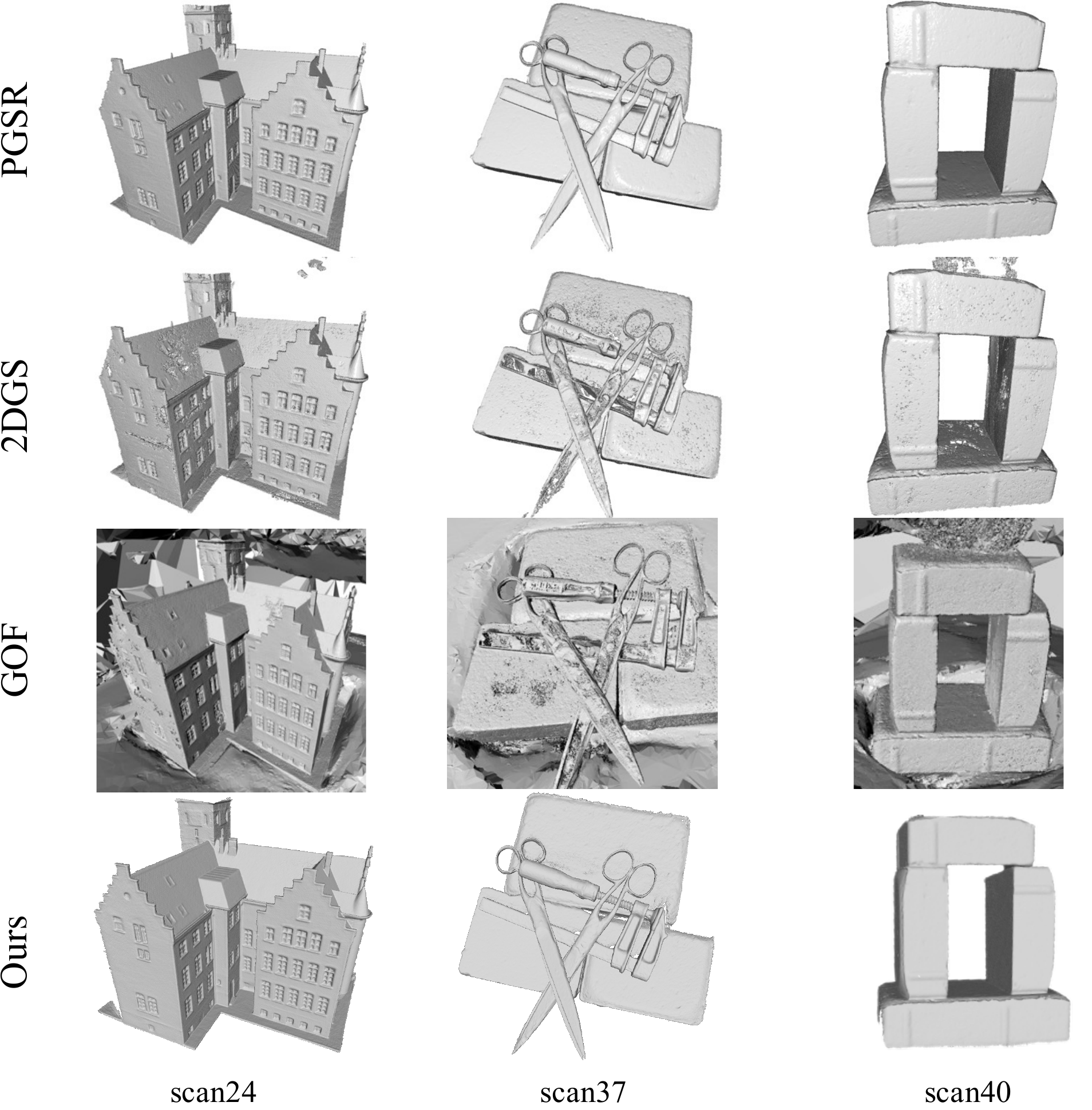}
	\vspace*{\fill} % 添加第二个垂直填充
	\caption{Additional qualitative reconstruction results on scenes from the DTU dataset. This demonstrates the effectiveness of our method (TSGS) on general opaque objects, achieving competitive reconstruction quality.}
	\label{fig:additional_qualitative_dtu_1} % Adjusted label
\end{figure*}

\begin{figure*}[htbp!]
	\centering
	\vspace*{\fill} % 添加第一个垂直填充
	\includegraphics[height=\mydtufigureheight, keepaspectratio=true]{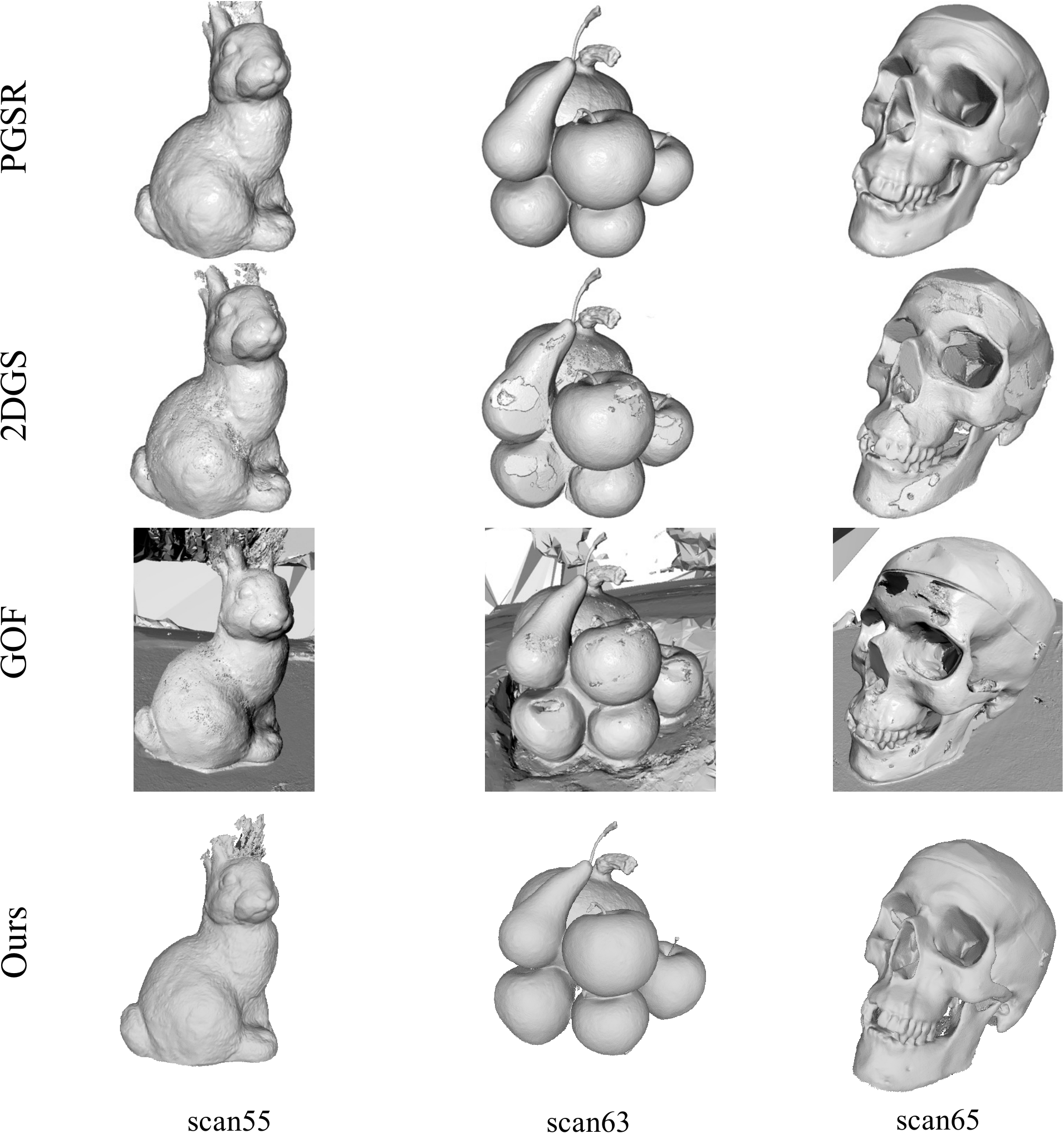}
	\vspace*{\fill} % 添加第二个垂直填充
	\caption{Additional qualitative reconstruction results on scenes from the DTU dataset (Continued). This demonstrates the effectiveness of our method (TSGS) on general opaque objects, achieving competitive reconstruction quality.} % Slightly modified caption
	\label{fig:additional_qualitative_dtu_2} % Adjusted label
\end{figure*}

\begin{figure*}[htbp!]
	\centering
	\vspace*{\fill} % 添加第一个垂直填充
	\includegraphics[height=\mydtufigureheight, keepaspectratio=true]{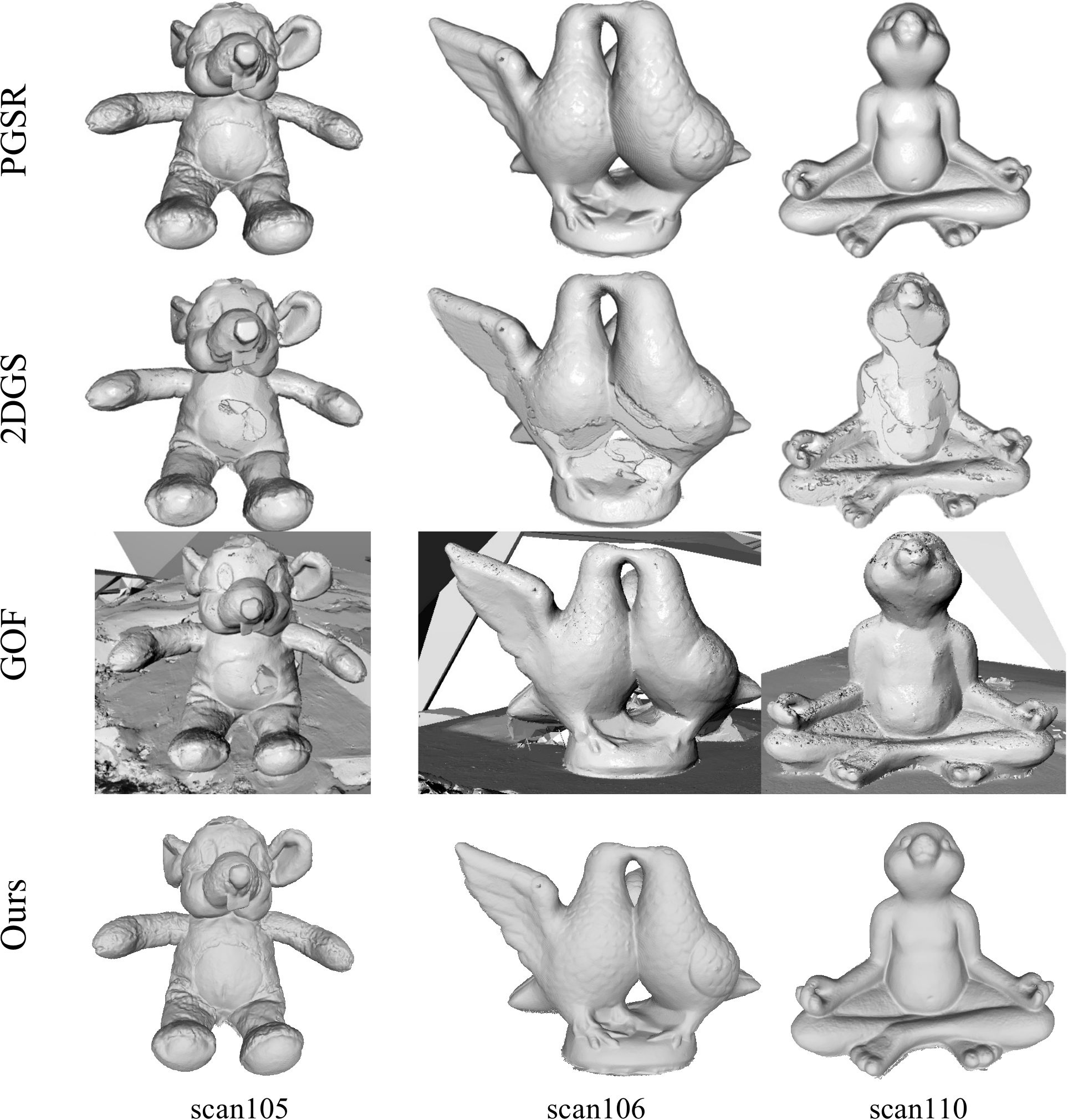}
	\vspace*{\fill} % 添加第二个垂直填充
	\caption{Additional qualitative reconstruction results on scenes from the DTU dataset (Continued). This demonstrates the effectiveness of our method (TSGS) on general opaque objects, achieving competitive reconstruction quality.} % Slightly modified caption
	\label{fig:additional_qualitative_dtu_3} % Adjusted label
\end{figure*}

\begin{figure*}[htbp!]
	\centering
	\vspace*{\fill} % 添加第一个垂直填充
	\includegraphics[height=\mydtufigureheight, keepaspectratio=true]{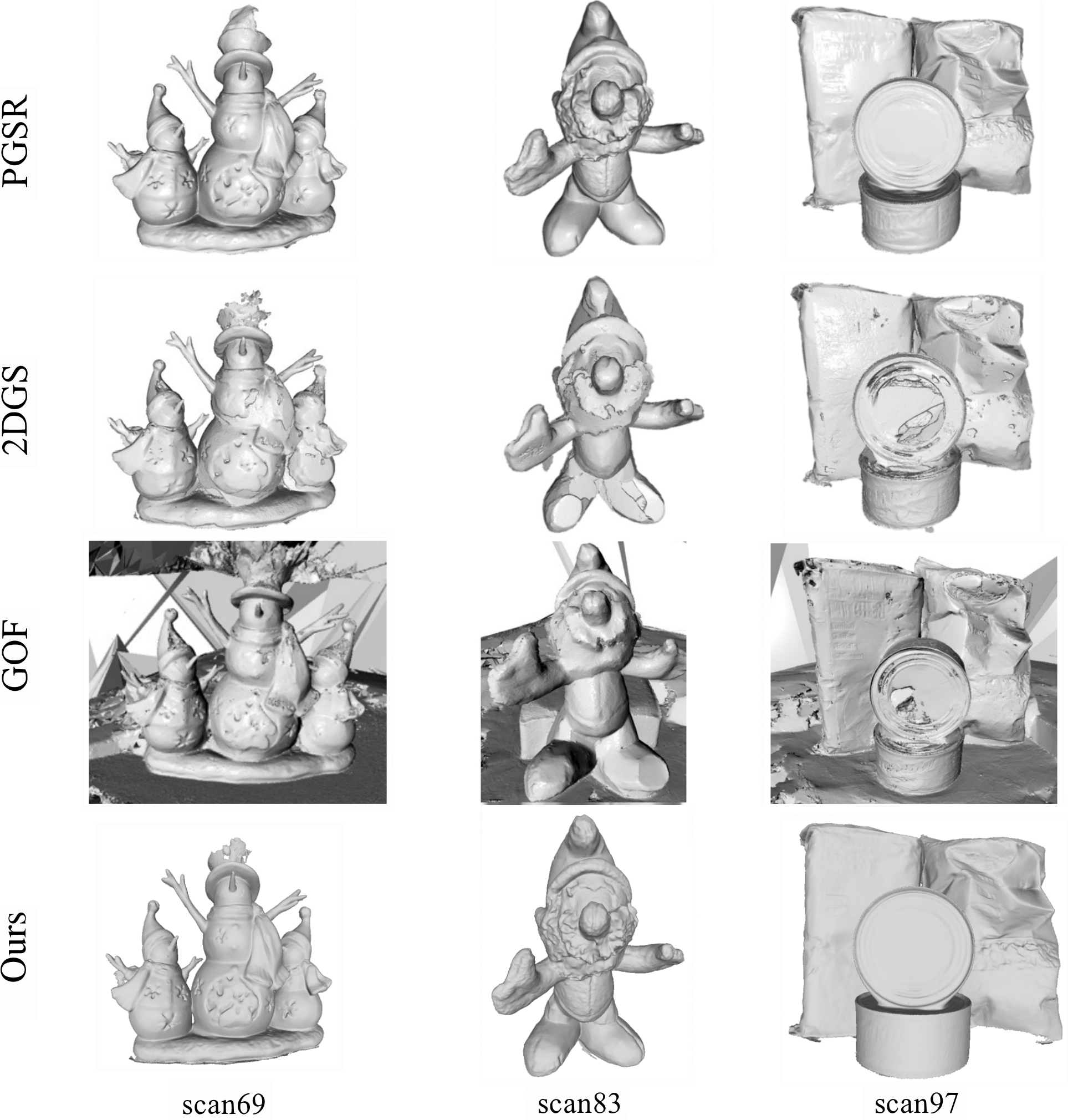}
	\vspace*{\fill} % 添加第二个垂直填充
	\caption{Additional qualitative reconstruction results on scenes from the DTU dataset (Continued). This demonstrates the effectiveness of our method (TSGS) on general opaque objects, achieving competitive reconstruction quality.} % Slightly modified caption
	\label{fig:additional_qualitative_dtu_4} % Adjusted label
\end{figure*}

\begin{figure*}[htbp!]
	\centering
	\vspace*{\fill} % 添加第一个垂直填充
	\includegraphics[height=\mydtufigureheight, keepaspectratio=true]{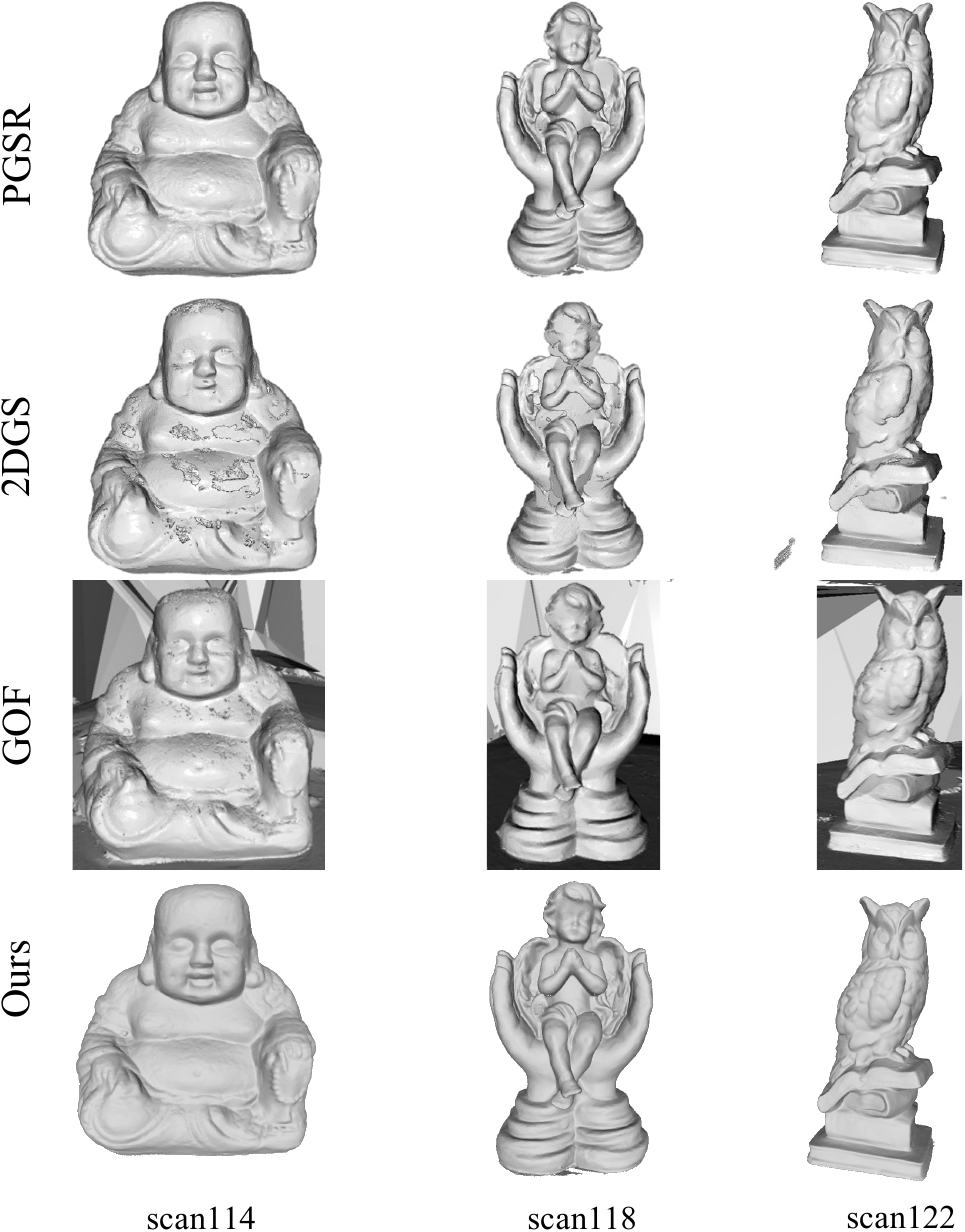}
	\vspace*{\fill} % 添加第二个垂直填充
	\caption{Additional qualitative reconstruction results on scenes from the DTU dataset (Continued). This demonstrates the effectiveness of our method (TSGS) on general opaque objects, achieving competitive reconstruction quality.} % Slightly modified caption
	\label{fig:additional_qualitative_dtu_5} % Adjusted label
\end{figure*}

\end{document}